\documentclass[sigconf]{acmart}
\usepackage[table]{xcolor}
\usepackage{enumitem}
\usepackage{amsmath}  

\usepackage{algorithm}  
\usepackage{algorithmic}  
\usepackage{multirow}  %
\definecolor{MyBlue}{HTML}{eaf1ef}
\definecolor{Myyellow}{HTML}{f8f4eb}
\definecolor{Mypink}{HTML}{f7eeeb}
\renewcommand\footnotetextcopyrightpermission[1]{}
\AtBeginDocument{%
  }

\setcopyright{acmlicensed}
\copyrightyear{2018}
\acmYear{2018}
\acmDOI{XXXXXXX.XXXXXXX}
\acmConference[Conference acronym 'XX]{Make sure to enter the correct
  conference title from your rights confirmation email}{June 03--05,
  2018}{Woodstock, NY}
\acmISBN{978-1-4503-XXXX-X/2018/06}

\begin{document}

\title{Mitigating Multimodal Hallucination via Phase-wise Self-reward}



\author{Yu Zhang}
\authornote{Both authors contributed equally to this research.}
\affiliation{%
  \institution{Harbin Institute of Technology}
  \institution{Peng Cheng Laboratory}
  \city{Shenzhen}
  \country{China}
}
\email{yuzhang2717@gmail.com}

\author{Chuyang Sun}
\authornotemark[1]
\affiliation{%
  \institution{Harbin Institute of Technology}
  \city{Shenzhen}
  \country{China}
}
\email{25s051016@stu.hit.edu.cn}

\author{Kehai Chen}
\affiliation{%
  \institution{Harbin Institute of Technology}
  \institution{Peng Cheng Laboratory}
  \city{Shenzhen}
  \country{China}
}
\email{chenkehai@hit.edu.cn}

\author{Xuefeng Bai}
\affiliation{%
  \institution{Harbin Institute of Technology}
  \city{Shenzhen}
  \country{China}
}
\email{baixuefeng@hit.edu.cn}



\author{Yang Xiang}
\affiliation{%
  \institution{Peng Cheng Laboratory}
  \city{Shenzhen}
  \country{China}}
\email{xiangy@pcl.ac.cn}

\author{Min Zhang}
\affiliation{%
  \institution{Harbin Institute of Technology}
  \institution{Peng Cheng Laboratory}
  \city{Shenzhen}
  \country{China}}
\email{zhangmin2021@hit.edu.cn}




\renewcommand{\shortauthors}{Trovato et al.}


\begin{abstract}
Large Vision-Language Models (LVLMs) still struggle with \textbf{vision hallucination}, where generated responses are inconsistent with the visual input. 
Existing methods either rely on large-scale annotated data for fine-tuning, which incurs massive computational overhead, or employ static post-hoc strategies that overlook the dynamic nature of hallucination emergence.
To address these, we introduce a new self-rewarding framework, enabling dynamic hallucination mitigation at inference time without external supervision.
On the empirical side, we reveal that visual hallucination exhibits phase-wise dynamic patterns, peaking at the onset of each semantic phase.
Drawing on these insights, we propose \textbf{\textsc{PSRD}} (\textbf{P}hase-wise \textbf{S}elf-\textbf{R}eward \textbf{D}ecoding) for online hallucination correction guided by phase-wise self-reward signals.
To reduce the cost of repeated self-evaluation during decoding, we distill the hallucination guidance signal from LVLMs into a lightweight reward model.
The reward model subsequently provides on-the-fly guidance for targeted intervention during the decoding process, enabling precise hallucination suppression.
The proposed \textsc{PSRD} significantly reduces the hallucination rate of LLaVA-1.5-7B by $50.0\%$ and consistently outperforms existing post-hoc methods across five hallucination evaluation benchmarks for four LVLMs.
Further analysis confirms that \textsc{PSRD} effectively mitigates hallucination propagation and achieves a highly controllable trade-off between strong performance and inference efficiency.
\end{abstract}

\received{20 February 2007}
\received[revised]{12 March 2009}
\received[accepted]{5 June 2009}

\maketitle
\section{Introduction}
\label{sec:intro}

Large Vision-Language Models (LVLMs) have achieved remarkable performance across diverse multimodal tasks~\cite{gpt4,qwen2.5vl,mme,mmmu,wei2026zooming,zhang2025cross,zhu2024benchmarking,zheng2025locot2v}.
However, they remain susceptible to \textbf{vision hallucinations}~\cite{bai2024hallucination,huang2024visual}, where generated content is factually inconsistent with the visual input. 
This limitation significantly hinders their reliability in real-world applications.
Existing mitigation strategies~\cite{ouali2024clip,xiao2025detecting,fu2025mitigating,yue2024less} primarily rely on large-scale human-annotated, or model-distilled preference data to fine-tune LVLMs through Direct Preference Optimization (DPO)~\cite{dpo}.
Despite their strong performance, these methods rely heavily on external supervision and full-model fine-tuning, which incur prohibitive costs.
\textbf{\begin{figure}[t]
    \centering
    \includegraphics[page=1, scale=0.47]{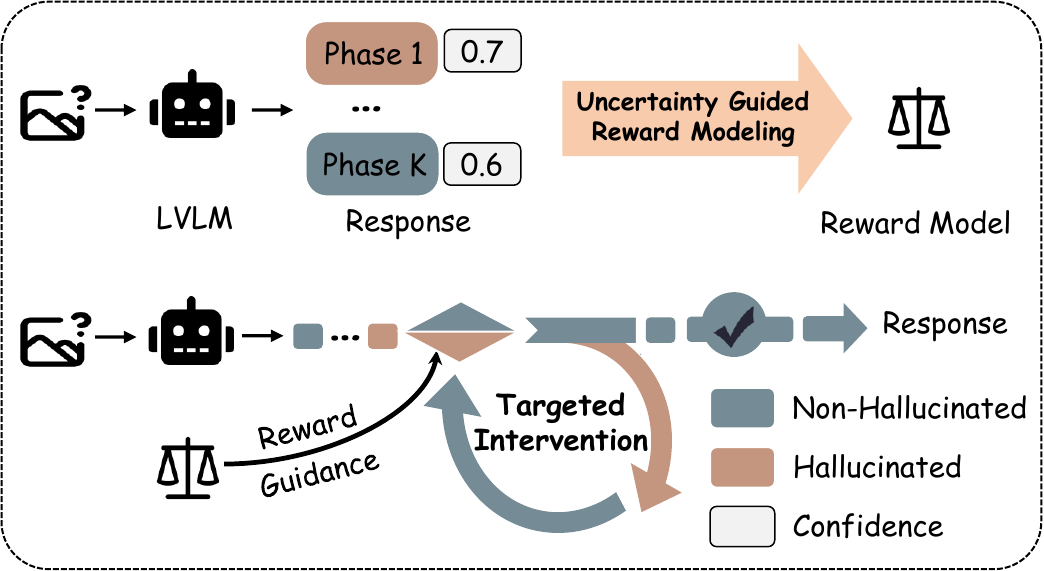}
    \caption{Illustration of the proposed \textsc{PSRD} framework. \textsc{PSRD} first activates the intrinsic hallucination discrimination capacity of LVLMs through the uncertainty signals to train a lightweight phase-wise reward model. 
    Then the reward model monitors the response online to provide on-the-fly reward signals, enabling dynamic, targeted intervention during the decoding process.
    }
    \label{fig:method___}
\end{figure}}
To circumvent these costs, recent efforts~\cite{leng2024mitigating,wang2024mitigating,manas2025controlling,wang2024mllm,yin2024woodpecker} have explored post-hoc mitigation strategies.
These include generate-then-revise methods~\cite{yin2024woodpecker,lee2023volcano} that perform a single correction pass after full response generation, and contrastive decoding methods~\cite{leng2024mitigating,wang2024mitigating} that apply continuous intervention at every decoding step by injecting contrastive signals.
However, these methods often overlook the \textit{dynamic nature} of hallucination emergence, leading to a lack of precise intervention at critical junctures.

\begin{figure}[h]
    \centering
    \includegraphics[width=0.95\linewidth]{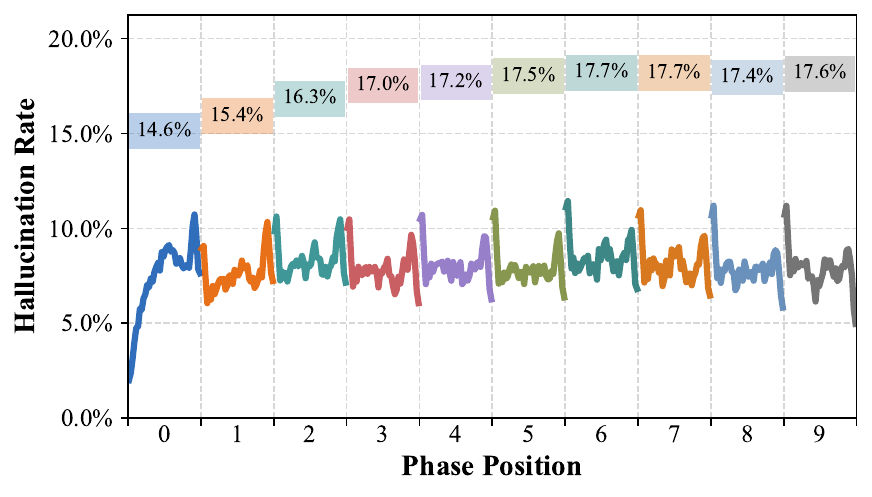}
    \caption{Characterization of dynamic hallucination patterns across and within generation phases. The upper panel illustrates the average hallucination rate across consecutive phases, while the lower panel depicts the word-level hallucination rate within each phase. The error rate peaks at the onset of each semantic segment, identifying these transitions as critical junctures for hallucination emergence.}
    \label{fig:feature}
\end{figure}

To address these issues, we introduce a new self-rewarding framework, which enables dynamic hallucination mitigation at inference time without external supervision.
We first conduct an in-depth analysis of the dynamic patterns of visual hallucination during decoding.
As shown in Figure \ref{fig:feature}, we reveal that vision hallucination exhibits distinct phase-wise patterns:
hallucination severity varies across different generation phases, notably peaking at the onset of each phase.
Motivated by these insights, we propose \textbf{\textsc{PSRD}} (\textbf{P}hase-wise \textbf{S}elf-\textbf{R}eward \textbf{D}ecoding)  for online hallucination correction guided by phase-wise self-reward signals, as shown in Figure~\ref{fig:method___}.
To ensure inference efficiency, we distill the LVLM’s latent discriminative capacity into a lightweight reward model. Specifically, we elicit this intrinsic capacity as phase-wise uncertainty signals, which are then utilized via an uncertainty-guided weighting mechanism to construct the reward model.
The reward model monitors the generation process online, providing on-the-fly reward signals to trigger iterative, targeted interventions at the early stages of problematic phases. 
In this way, \textsc{PSRD} achieves dynamic, on-the-fly hallucination suppression, circumventing the need for expensive external annotations or computationally intensive fine-tuning of the underlying LVLM.

Extensive experiments across five benchmarks and three distinct LVLMs demonstrate the superior effectiveness of our framework. 
Specifically, \textsc{PSRD} reduces hallucinations in LLaVA-1.5-7B by $50.0\%$ and consistently outperforms existing post-hoc methods across four LVLMs. 
Quantitative analysis also reveals that by targeting and suppressing hallucinations at the onset of each phase, \textsc{PSRD} significantly mitigates hallucination propagation. 
This is evidenced by a phase-level hallucination accumulation rate of only $0.07\%$, which is approximately seven times lower than that of the base model. 
Furthermore, our analysis confirms that \textsc{PSRD} enables a highly controllable trade-off between the performance of hallucination mitigation and inference efficiency.

Our main contributions are summarized as follows:

\begin{itemize}[leftmargin=1.2em]
    \item \textit{We reveal that visual hallucination exhibits phase-wise dynamic patterns, peaking at the onset of each semantic phase.}
    \item \textit{We propose \textsc{PSRD}, a self-rewarding framework that enables dynamic, phase-wise correction at inference time.}
    \item \textit{Extensive experiments across five benchmarks and three LVLMs demonstrate that \textsc{PSRD} consistently outperforms existing post-hoc hallucination mitigation methods. }
    \item \textit{Further analysis confirms that \textsc{PSRD} effectively mitigates hallucination propagation and achieves a controllable trade-off between strong performance and considerable inference efficiency.}
\end{itemize}

\section{Analysis of Visual Hallucination Dynamics}
\label{sec:Dynamics}

In this section, we conduct quantitative experiments to reveal that multimodal hallucinations exhibit distinct phase-wise characteristics during the decoding process. 
We randomly sampled 500 images from the COCO2014 dataset~\cite{lin2014microsoft} and generated captions using LLaVA-1.5-7B~\cite{llava1.5}. 
Following the evaluation protocol in \citep{rohrbach2018object}, we utilize both phase-level and word-level hallucination rates to quantify severity across different stages of decoding.
We define a phase as a fine-grained, semantically coherent unit obtained by segmenting captions using predefined textual delimiters. 
This granular definition allows for a more localized assessment of the model's behavior throughout the generation trajectory. 
The specific procedures for calculating hallucination rates are detailed in Appendix \ref{supp:metric_hall}.

As illustrated in Figure \ref{fig:feature}, hallucination severity exhibits a distinct phase-wise dynamic, consistently peaking at the onset of each segment.
While prior work~\cite{early} has identified variations across different phases, our analysis uncovers a more granular intra-phase pattern. 
This recurring pattern suggests that the transition into a new semantic segment represents a vulnerable critical juncture. 
At these points, the model must initiate a new descriptive element while re-aligning its evolving linguistic context with the visual evidence—a process that is highly susceptible to grounding failures. 
Once this initial alignment is established, subsequent tokens within the same phase benefit from more stable contextual grounding, leading to a marked reduction in hallucination frequency. 
These empirical observations directly motivate the design of a phase-aware decoding framework, which enables targeted intervention at these specific vulnerable junctures to preemptively suppress hallucination emergence.

\section{Phase-wise Self-Reward Decoding}
Building upon the empirical findings, we propose \textsc{PSRD} (Phase-wise Self-Reward Decoding) for on-the-fly multimodal hallucination correction with targeted intervention guided by phase-wise self-reward.
The essence of such a dynamic system lies in its ability to provide precise, real-time guidance at each semantic phase of the decoding process. 
However, directly querying the LVLMs for iterative assessment is too slow to support on-the-fly guidance.
Therefore, we distill the LVLMs’ latent hallucination discrimination capacity into a lightweight re-calibrated reward model with an uncertainty-guided weighting mechanism (\S \ref{sec:threshold}).
By integrating phase-wise self-reward signals with a targeted intervention strategy, PSRD achieves precise hallucination suppression during decoding (\S \ref{sec:mitigation}).

\subsection{Uncertainty-guided Reward Model Construction.}
\label{sec:threshold}
We leverage self-elicited phase-wise hallucination behaviors as uncertainty signals and utilize an uncertainty-guided weighting mechanism to construct the lightweight re-calibrated reward model.

\paragraph{\textbf{Hallucination Behavior Elicitation.}}
\label{sec:generation}
We elicit a broad spectrum of hallucination behaviors by collecting model responses under controlled perturbations of visual and textual inputs. Specifically, we employ two types of visual inputs—the original image and a noise-corrupted version—paired with two distinct textual instructions: a standard prompt and a hallucination-inducing prompt designed to provoke speculative descriptions. These four configurations are used to systematically enrich the frequency and semantic diversity of hallucination behaviors in the collected responses. Detailed prompt templates and data generation procedures are provided in Appendix~\ref{supp:generation}.

\paragraph{\textbf{Phase-wise Uncertainty Signals Construction.}}

Inspired by the empirical analysis of hallucination dynamics (\S \ref{sec:Dynamics}), we first segment each generated response $\hat{y}$ into a sequence of fine-grained, semantically coherent phrases $\{s_1, s_2, \dots, s_n\}$ that serve as distinct generation phases. 
Then we leverage the LVLM’s intrinsic self-evaluation capacity to obtain uncertainty supervision signals. 
For each phrase $s_i$, we pair it with the corresponding clean image $I$ and re-prompt the LVLM $\mathcal{M}$ to determine its factual consistency with the image.
Beyond the binary judgment, we extract the LVLMs' internal confidence scores as raw uncertainty signals.
Formally, the uncertainty signal for phase $s_i$ is captured by the softmax probabilities $(p_i^+, p_i^-) \in [0,1]$ corresponding to "Grounded" ($\text{NonH}$) and "Hallucinated" ($\text{H}$) labels:
\begin{equation}
(p_i^+, p_i^-) =
\left[ \mathrm{Softmax}(\mathbf{W}^\top \mathbf{h}_{s_i}) \right]_{[\text{NonH}, \text{H}],} 
\end{equation}
where $\mathbf{h}_{s_i}$ denotes the hidden representation of the final token in $s_i$ and $\mathbf{W}$ is the unembedding matrix of the LVLM. Additional details on phase segmentation and the self-evaluation prompt are provided in Appendix~\ref{supp:generation}.

\paragraph{\textbf{Uncertainty-Guided Reward Calibration.}}
Leveraging these phase-wise weak supervision signals ($(p_i^+, p_i^-)$), we develop the re-calibrated reward model ($\mathcal{R}$) through an uncertainty-guided weighting mechanism.
We initialize $\mathcal{R}$ with a CLIP backbone~\cite{ouali2024clip}, utilizing its discriminative, non-sequential architecture.
For each elicited triplet $(I, s_k^+, s_k^-) \in \mathcal{D}$ consisting of an image and its corresponding grounded and hallucinatory phrases, we extract the image embedding $\mathbf{v}_I$ and text embeddings $\mathbf{t}_{I,k}^+$ and $\mathbf{t}_{I,k}^-$ from the final-layer latent representations of the CLIP vision and text encoders, respectively.
Following standard CLIP conventions, the reward value—representing the degree of visual-semantic alignment—is defined as the cosine similarity between the image and text representations:
\begin{equation}
\label{eq:SCORE}
\mathbf{c}_{I,k}^+ = \cos(\mathbf{v}_I, \mathbf{t}_{I,k}^+)\ , 
\mathbf{c}_{I,k}^- = \cos(\mathbf{v}_I, \mathbf{t}_{I,k}^-).
\end{equation}

To optimize the reward model $\mathcal{R}$ for robust feature separability and semantic consistency, we employ a multi-objective framework modulated by an uncertainty-guided weighting mechanism. 
Specifically, we utilize the product of uncertainty signals $p_{I,k}^+ p_{I,k}^-$ as a joint modulation factor to filter the supervision signals. 
We first adopt a Discriminative Alignment (DA) Loss to supervise the model in distinguishing between grounded and hallucination-prone phrases.
Framed as a binary classification task, this objective maximizes the alignment score of non-hallucinated phrases relative to the hallucinated counterparts, thereby enforcing stricter semantic grounding:
\begin{equation}
\mathcal{L}_{\text{DA}} = \mathbb{E}_{(I, k) \sim \mathcal{D}} \left[\ell_{\text{CE}} \left( \left[ \mathbf{c}_{I,k}^-,\ \mathbf{c}_{I,k}^+ \right],\ \mathbf{1} \right) \cdot p^+_{I,k}p^-_{I,k} \right].
\end{equation}
Here, $\ell_{\text{CE}}(\cdot)$ is the cross-entropy loss, $\mathbf{1}$ denotes the index for the positive (non-hallucination) class.
To further promote more robust representation separation in the feature space, we introduce Margin Enforcement Loss by explicitly enforcing a minimum margin $\delta$ between the positive and negative alignment scores:
\begin{equation}
\mathcal{L}_{\text{Margin}} = \mathbb{E}_{(I, k) \sim \mathcal{D}} \left[\max(0,\ \mathbf{c}_{I,k}^- - \mathbf{c}_{I,k}^+ + \delta) \cdot p_{I,k}^+ p_{I,k}^- \right].
\end{equation}

Additionally, Hallucination Consistency (HC) Loss is used to promote invariance in learned representations against the diversity of negative samples by encouraging all hallucinated sentences $(s_k^-, s_{k'}^-)$ generated for the same image $I$ to cluster closely:
\begin{equation}
\mathcal{L}_{\text{HC}} = \mathbb{E}_{(I, s_k^-, s_{k'}^-) \sim \mathcal{D}} \left[ (1-\text{cos}(\mathbf{t}_{I,k}^-, \mathbf{t}_{I,k'}^-)) \cdot p_{I,k}^- p_{I,k'}^- \right].
\end{equation}
Finally, the objective function is as follows:
\begin{equation}
\label{eq:loss_total}
\mathcal{L}_{\text{total}} = \lambda_1 \mathcal{L}_{\text{DA}} + \lambda_2 \mathcal{L}_{\text{Margin}} + \lambda_3 \mathcal{L}_{\text{HC}},
\end{equation}
where $\lambda_1$, $\lambda_2$, and $\lambda_3$ balance the contribution of different learning objectives.
In this way, we construct a lightweight yet robust reward model $\mathcal{R}$ capable of providing precise, phase-wise guidance during the LVLM decoding process.

\begin{table*}[t]
\centering
\renewcommand\arraystretch{1.2}
\begin{minipage}{0.5\linewidth}
    \centering
\caption{Comparison with the state-of-the-art methods for the generative hallucination mitigation tasks across three benchmarks.
\textbf{Bold} indicates the best result among post-hoc methods, and \underline{underline} indicates the second-best.}
  \scalebox{0.8}
  {`
    \begin{tabular}{l c cccc c}
    \toprule
      
      \multirow{2}{*}{\textbf{Method}}   & \multicolumn{4}{c}{\textbf{AMBER}} & \multicolumn{2}{c}{\textbf{MMHal-Bench}}  \\

      \cmidrule(lr){2-5} \cmidrule(lr){6-7}
      
       & CHAIR $\downarrow$ & Cover $\uparrow$ & Hal $\downarrow$ &  Cog $\downarrow$ & Overall $\uparrow$ & Hal $\downarrow$   \\
    \midrule
      LLaVA-1.5-7B (Base)      & 7.8    & 51.0   & 36.4    & 4.2 &  1.55  & 0.76   \\
      GPT-4V        & 4.6  & 67.1 & 30.7 &  2.6  & 3.49 & 0.28 \\
      \midrule
    
    \rowcolor{Mypink}
    
    \multicolumn{7}{c}{\textbf{\textit{Fine-tune LVLMs with external annotation data}}}  \\ 
    \text{EOS} (\texttt{\scriptsize{ACL'24}}) &5.1   & 49.1  &  22.7 &  2.0&2.03&0.59\\
    \text{LLaVA-DPO} (\texttt{\scriptsize{ICLR'25}}) & 2.8 & 47.8 & 15.5 & 1.6 & 2.58 & 0.50 \\
    \text{HSA-DPO} (\texttt{\scriptsize{AAAI'25}}) & 2.1 & 47.3 &  13.4  & 1.2  & 2.61  & 0.48 \\
    \text{CLIP-DPO} (\texttt{\scriptsize{ECCV'24}}) & 3.7 & 47.8 & 16.6 & 1.3 &-&-\\
    \text{RLAIF-V} (\texttt{\scriptsize{CVPR'25}}) & 2.9 & 50.2 & 16.0& 1.0&  -& - \\  
        \midrule
    \rowcolor{Myyellow} 

              \multicolumn{7}{c}{\textbf{\textit{Fine-tune LVLMs with self-improvement}}}  \\ 
    \text{STIC} (\texttt{\scriptsize{NeurIPS'25}})  &7.6  & 52.1 & 35.8&4.4 &2.07  & 0.56 \\     
    \text{SENA} (\texttt{\scriptsize{AAAI'25}})  & 4.9 &49.4  & 20.5 & 1.7 & 2.33 & 0.52 \\    
      \midrule
          \rowcolor{MyBlue} 
                  \multicolumn{7}{c}{\textbf{\textit{Post-hoc hallucination mitigation} (LLaVA-1.5-7B)}}  \\ 
      VCD (\texttt{\scriptsize{CVPR'24}})  & 6.7 & 46.5 & 27.8 & 2.0 & 1.96 & 0.64 \\
      M3ID (\texttt{\scriptsize{CVPR'24}}) & 6.0 & 48.9 & 26.0 & \underline{1.5} & 2.14 & 0.61 \\
      OPERA (\texttt{\scriptsize{CVPR'24}}) & -& -&- & -&2.15 & 0.54 \\
      AVISC (\texttt{\scriptsize{ACL'25}}) &  6.3 & 46.6 &  25.6 &  2.0 &  2.19 & 0.59 \\
      MoD (\texttt{\scriptsize{ACL'25}}) &  8.6 &  \underline{50.7}& 38.8  &  4.7 &  - &-  \\
      MRGD (\texttt{\scriptsize{ICCV'25}}) &  \underline{4.4} &  \textbf{62.5}& 21.9  & - &  - &-  \\
      Octopus (\texttt{\scriptsize{CVPR'25}})  & 4.8 &49.2 & \underline{23.4} &\textbf{1.2} & \underline{2.61} & \underline{0.50}\\
    \midrule
        \textbf{\textsc{PSRD}} (Ours)  &\textbf{3.9} & 48.2 & \textbf{20.1} & 2.0 &  \textbf{2.92} & \textbf{0.49}  \\ 
   \bottomrule
    \end{tabular}
  }

\label{tab:main1}
\end{minipage}
\hfill
\begin{minipage}{0.46\linewidth}
    \centering
\caption{Comparison with the state-of-the-art methods for the generative hallucination mitigation tasks across three benchmarks.
\textbf{Bold} indicates the best result among post-hoc methods, and \underline{underline} indicates the second-best.}
  \scalebox{0.81}
  {
    \begin{tabular}{lc c}
    \toprule
    \multirow{2}{*}{\textbf{Method}}   & \multicolumn{2}{c}{\textbf{Object HalBench} }  \\

      \cmidrule(lr){2-3} 
    & $\text{CHAIR}_{S} $$\downarrow$ & $\text{CHAIR}_{I}$ $\downarrow$   \\
    \midrule
      LLaVA-1.5-7B (Base)     & 46.3    & 22.6   \\

      GPT-4V       & 13.6 & 7.3  \\
      \midrule
    \rowcolor{Mypink}
    \multicolumn{3}{c}{\textbf{\textit{Fine-tune LVLMs with external annotation data}}}  \\ 
    \text{LLaVA-RLHF} (\texttt{\scriptsize{ACL'24}}) &  38.1&  18.9   \\
\text{RLAIF-V} (\texttt{\scriptsize{CVPR'25}}) & 10.5 & 5.2 \\  
    \text{HDPO} (\texttt{\scriptsize{ACL'25}}) & 16.6& 5.1 \\
    \text{EOS} (\texttt{\scriptsize{ACL'24}}) &40.3 & 17.8\\
        \midrule
          \rowcolor{MyBlue} 
                  \multicolumn{3}{c}{\textbf{\textit{Post-hoc hallucination mitigation}}}  \\ 
      VCD (\texttt{\scriptsize{CVPR'24}})  & 48.8 & 24.3  \\
      ICD (\texttt{\scriptsize{ACL'24}}) &47.4& 13.9 \\
      HALC (\texttt{\scriptsize{ICML'24}}) & 21.7 & 7.1  \\ 
      M3ID (\texttt{\scriptsize{CVPR'24}}) & 47.8 & 12.5  \\
      OPERA (\texttt{\scriptsize{CVPR'24}}) & 45.1 & 22.3 \\
      AVISC (\texttt{\scriptsize{ACL'25}}) &  49.6 & 12.9 \\
      MoD (\texttt{\scriptsize{ACL'25}}) &42.6   & 12.4  \\
      DeCo (\texttt{\scriptsize{ICLR'25}}) & 37.8 & 11.1  \\ 
      EAZY (\texttt{\scriptsize{ICCV'25}}) &38.8   & 11.4  \\
      ONLY (\texttt{\scriptsize{ICCV'25}}) &20.0   & 6.2 \\
      Octopus (\texttt{\scriptsize{CVPR'25}})   &  \underline{20.8} & \underline{6.6} \\
    \midrule
        \textbf{\textsc{PSRD}} (Ours) &  \textbf{10.1} & \textbf{4.1}   \\ 

   \bottomrule
    \end{tabular}
  }
\label{tab:main2}
\end{minipage}
\end{table*}

\subsection{Reward-guided Targeted Intervention}
\label{sec:mitigation}

To mitigate hallucinations while keeping the search overhead manageable, we formulate decoding intervention as a constrained search problem. Let $\mathcal{S} = \{(k, \alpha) \mid k \in \mathbb{N}_{\ge 0}, \alpha \in \mathbb{R}_{\ge 0}\}$ denote the search space, where $k$ denotes the rank of an alternative initial token used to perturb the decoding trajectory and $\alpha$ denotes the contrastive penalty weight controlling the intervention strength.

The reward model is formulated as $\mathcal{R}: \mathcal{S} \to \mathbb{R}$, which evaluates the factual consistency of the generated phrase. Our objective is to find a satisficing solution $\mathbf{x}^* = (k^*, \alpha^*)$ such that $\mathcal{R}(\mathbf{x}^*) > \tau$ with a small number of reward model evaluations $N_{\mathrm{eval}}$, where $\tau$ is a pre-defined acceptance threshold.
The search space over $(k,\alpha)$ has a mixed structure. We treat the initial token rank $k$ as a discrete variable, since different $k$ values may induce qualitatively different decoding trajectories and thus very different rewards $\mathcal{R}(k,\cdot)$. In contrast, for a fixed seed trajectory, varying $\alpha$ often produces a usable local reward trend within a bounded probing interval, although the reward landscape in discrete decoding is generally neither globally smooth nor globally monotonic. Therefore, PSRD does not rely on any global monotonicity assumption over the full decoding space. Instead, we use a two-stage \textit{Scout-and-Project} strategy: we first identify a promising seed trajectory by low-cost discrete scouting, and then perform bounded local projection over $\alpha$ with fallback to the best observed candidate whenever the projected update is unreliable.

\paragraph{\textbf{Stage 1: Low-Cost Seed Scouting.}}
We first perform a warm-up scan over the top-$K$ most probable tokens with zero intervention ($\alpha=0$), which serves to identify a high-potential seed trajectory. We select the candidate $k^*$ that maximizes the initial reward:
$$
    k^* = \operatorname*{arg\,max}_{k \in \{0, \dots, K-1\}} \mathcal{R}(k, 0).
$$
If $\mathcal{R}(k^*, 0) > \tau$, the search terminates immediately.

\paragraph{\textbf{Stage 2: Bounded Local Projection.}}
Given the selected seed $k^*$, we refine the intervention strength $\alpha$ within a bounded interval rather than solving for a globally valid root. We first evaluate a small probe step $\delta$ to estimate a local finite-difference trend:

$$
    m \approx \frac{\mathcal{R}(k^*, \delta) - \mathcal{R}(k^*, 0)}{\delta}.
$$
This quantity is used only as a local heuristic for proposing the next intervention strength.
$$
    \alpha_{next} = \delta + \eta \cdot \frac{\tau - \mathcal{R}(k^*, \delta)}{m},
$$
where $\eta \ge 1.0$ is a relaxation factor. The projected value is clipped to a valid bounded interval before evaluation. If the estimated slope is unstable, degenerate, or yields an implausible update, we do not continue an unconstrained secant iteration; instead, we fall back to the best candidate observed in the probed interval. In this way, the projection step serves as a search heuristic for quickly locating a satisfactory intervention strength, while preserving robustness when the local reward trend is weak or inconsistent.

The full procedure is given in Algorithm~\ref{alg:search}. Implementation details are provided in Appendix~\ref{sec:InterAlgorithm}, and additional analyses on local reward trends, fallback behavior, and intervention-strength trade-offs are provided in Appendix~\ref{app:monotonicity_fallback} and ~\ref{app:fluency_alpha}.

\begin{table*}[t]
\centering
\small
\caption{Comparison with the state-of-the-art methods for the discriminative hallucination mitigation tasks across two benchmarks. } 
\resizebox{0.95\textwidth}{!}{
\begin{tabular}{lcccccccccc}
\toprule
\multirow{3}{*}{\textbf{Method}}
 &\multicolumn{2}{c}{\textbf{AMBER}}&\multicolumn{8}{c}{\textbf{POPE}} \\ 
 \cmidrule[0.4pt](lr){2-3} 
 \cmidrule[0.4pt](lr){4-11}
 &\multicolumn{2}{c}{Discrimination}&\multicolumn{2}{c}{Random}&\multicolumn{2}{c}{Popular}&\multicolumn{2}{c}{Adversarial}&\multicolumn{2}{c}{ALL}  \\
 \cmidrule[0.4pt](lr){2-3}
 \cmidrule[0.4pt](lr){4-5}
 \cmidrule[0.4pt](lr){6-7}
 \cmidrule[0.4pt](lr){8-9}
 \cmidrule[0.4pt](lr){10-11}
 &Accuracy &F1 &Accuracy &F1 &Accuracy &F1 &Accuracy &F1 &Accuracy &F1\\
 \midrule
  LLaVA-1.5-7B (Base)
  & 67.0 & 71.1& 83.8  & 81.9  & 82.6  & 80.9  & 79.8  & 78.5  & 82.0  & 80.4 \\
  +OPERA (\texttt{\scriptsize{CVPR'24}})& - &- &84.4  &-   &83.4   &-   &81.2  &-  &83.0  &-   \\
 +VCD (\texttt{\scriptsize{CVPR'24}})
 & 67.3 & 71.1& 85.4 & 84.0  & 83.2  & 81.9  & 80.3 & 79.5  & 83.0  & 81.8 \\
 +M3ID (\texttt{\scriptsize{CVPR'24}}) 
 & 67.3& 70.9& 86.1  & 81.9  & 82.1  & 80.8  & 79.5  & 78.2  & 82.6  & 80.3  \\
  +ConVis (\texttt{\scriptsize{AAAI'25}})& - &- &84.7  &-   &83.2   &-   &81.1  &-  &83.0  &-   \\
  +AVISC (\texttt{\scriptsize{ACL'25}})
  & 70.7& 75.5& 84.7  & 82.2  & 83.7  & 81.3  &\underline{81.8}  & 79.6  & 83.4  & 81.0  \\
 +ONLY (\texttt{\scriptsize{ICCV'25}}) & -&-&\textbf{89.7}  & \underline{89.1}  &\underline{86.0}& \textbf{86.3}& 79.4 & 81.1  &85.0  & 85.5 \\
  +EAZY (\texttt{\scriptsize{ICCV'25}}) & -&-&  - &- &- &- & - &-  &85.0  & \underline{85.8} \\
  +ALGA (\texttt{\scriptsize{CVPR'25}}) & {-} & {-}& \underline{88.5}& 87.7& 85.1& 84.7& {81.1}& \underline{81.4}  & {84.9}  & 84.6 \\
 +Octopus (\texttt{\scriptsize{CVPR'25}}) & \underline{76.7}& \underline{82.7}& 87.5  & 85.4  & \underline{85.2}& 84.2 & \textbf{82.2}  & \underline{81.4}  &\underline{85.8}  & 83.4 \\
  \textbf{+\textsc{PSRD}} (Ours) & \textbf{81.2} & \textbf{85.0}& \textbf{89.7}  & \textbf{89.8}  & \textbf{86.3}  & \underline{85.8}  & \textbf{82.2}  & \textbf{82.3}  & \textbf{86.1}  & \textbf{86.0} \\
\bottomrule
\end{tabular}
}
\label{tab:table2}

\end{table*}

\section{Experiments}
\label{sec:experiments}

\subsection{Experiment Settings}
In this section, we briefly introduce the experimental settings, with full details in Appendix~\ref{supp:exp}.

\paragraph{\textbf{Benchmarks \& Metrics}}
We conduct generative and discriminative hallucination mitigation experiments following established evaluation protocols~\cite{amber,suo2025octopus}. 
For generative tasks, we evaluate \textit{Object HalBench}~\cite{rohrbach2018object} ($\text{CHAIR}_i$, $\text{CHAIR}_s$), \textit{AMBER}~\cite{amber} (CHAIR, Cover, Hal and Cog), and \textit{MMHal-Bench}~\cite{llava-rlhf} (Overall and Hal). 
For discriminative tasks, we evaluate \textit{POPE}~\cite{pope} and \textit{AMBER}~\cite{amber} using Accuracy and F1-score. Besides, we validate the performance of the reward model on sentence-level hallucination classification tasks across \textit{AMBER HalDet}~\cite{amber} and \textit{MHal-detect}~\cite{gunjal2024detecting}, measured by  Accuracy, Precision, Recall, and F1-score.

\paragraph{\textbf{Baselines}}
For hallucination mitigation, we compare existing approaches from four categories: 
(1) \textit{Standard LVLMs}, including LLaVA-1.5-7B~\cite{llava1.5}; 
(2) \textit{Fine-tuned LVLMs with externally annotated data}, represented by HDPO~\cite{fu2025mitigating}; 
(3) \textit{Fine-tuned LVLMs via self-improvement}, exemplified by STIC~\cite{deng2024enhancing}; and 
(4) \textit{Post-hoc methods}, for which we adopt M3ID~\cite{favero2024multi}, a decoding-time approach based on dynamic intervention.
For completeness, detailed descriptions of the remaining baselines are provided in Appendix~\ref{sec:appendix-baselines}.

\begin{table}[t]
    \centering
    \small
    \caption{Results on the AMBER generative benchmark for LLaVA-Next-7B, InstructBLIP-7B, LLaVA-1.5-13B.}
    \resizebox{\linewidth}{!}{
    \begin{tabular}{lcccccccc}
        \toprule
        \multirow{2}{*}{\textbf{Method}} & \multicolumn{4}{c}{\textbf{AMBER}}  \\
        \cmidrule(lr){2-5} 
        & CHAIR $\downarrow$ & Cover $\uparrow$ & Hal $\downarrow$ &  Cog $\downarrow$  \\
        \midrule
        InstructBLIP-7B  &8.4 & 46.4 & 31.1 & 2.6  \\
        +M3ID (\texttt{\scriptsize{CVPR'24}}) & 6.9 & 47.2 & 27.5 & 2.2\\
        +VCD (\texttt{\scriptsize{CVPR'24}}) & 7.6 & 47.7 & 29.9 & 2.2 \\
        +AVISC (\texttt{\scriptsize{ACL'25}}) & 6.7 & 46.7 & 28.0 & 2.6 \\
        +Octopus (\texttt{\scriptsize{CVPR'25}}) & \underline{6.1} & \textbf{48.5} & \underline{22.2} & \textbf{1.3} \\
        \textbf{+\textsc{PSRD}} (Ours) &  \textbf{4.4}& \underline{47.8} &\textbf{20.9} & \underline{1.8}  \\ 
        \midrule
        LLaVA-Next-7B  & 7.1&55.8  &37.6 & 3.1 \\
            +VCD (\texttt{\scriptsize{CVPR'24}}) & 7.1 & \textbf{56.1} & 36.3  &  3.0 \\
            +AVISC (\texttt{\scriptsize{ACL'25}})& 7.5 & 55.2 & 38.0  &  \underline{2.5} \\
        +M3ID (\texttt{\scriptsize{CVPR'24}})& \underline{6.9} & \underline{55.6} & \underline{35.8}  &  3.3 \\
        \textbf{+\textsc{PSRD}} (Ours) & \textbf{4.7} & 45.3 & \textbf{21.1} & \textbf{1.2} \\
                \midrule
        LLaVA-1.5-13B  &  6.7    & 49.4   & \underline{28.8}    & \underline{3.1}   \\
            +VCD (\texttt{\scriptsize{CVPR'24}}) & 6.7 &  48.3 &  33.6 &  3.5 \\
            +AVISC (\texttt{\scriptsize{ACL'25}})& 8.6 & \underline{49.5} & 39.6  & 3.6  \\
        +M3ID (\texttt{\scriptsize{CVPR'24}})& \underline{6.0} & 47.7 & 30.4  &  3.3 \\
        \textbf{+\textsc{PSRD}} (Ours) & \textbf{4.7} & \textbf{52.0} & \textbf{24.1} &  \textbf{2.2}  \\

        \bottomrule
    \end{tabular}
    }

    \label{tab:other_model}

\end{table}
\subsection{Main Results}
The results of generative hallucination mitigation are in Table~\ref{tab:main1} and ~\ref{tab:main2}. 
\textsc{PSRD} significantly reduces the hallucination of LLaVA-1.5-7B by $50.0\%$, as measured by the CHAIR metric on AMBER and outperforms existing post-hoc methods across all metrics on the MMHal-Bench and Object HalBench.
For the AMBER dataset, \textsc{PSRD} achieves lower hallucination compared to baselines measured by CHAIR, Hal, and Cog metrics, while maintaining a high object cover rate. 

Notably, our method operates without fine-tuning base LVLMs or requiring external annotated data; the reliability of the self-evaluation signals used as weak supervision is further analyzed in Appendix~\ref{app:pseudo_label_reliability}.
\textsc{PSRD} outperforms existing self-improvement methods that rely on fine-tuning base LVLMs with self-supervision signals on AMBER and MMHal-Bench datasets. 
\begin{figure}[t]
\centering
\includegraphics[width=1\linewidth]{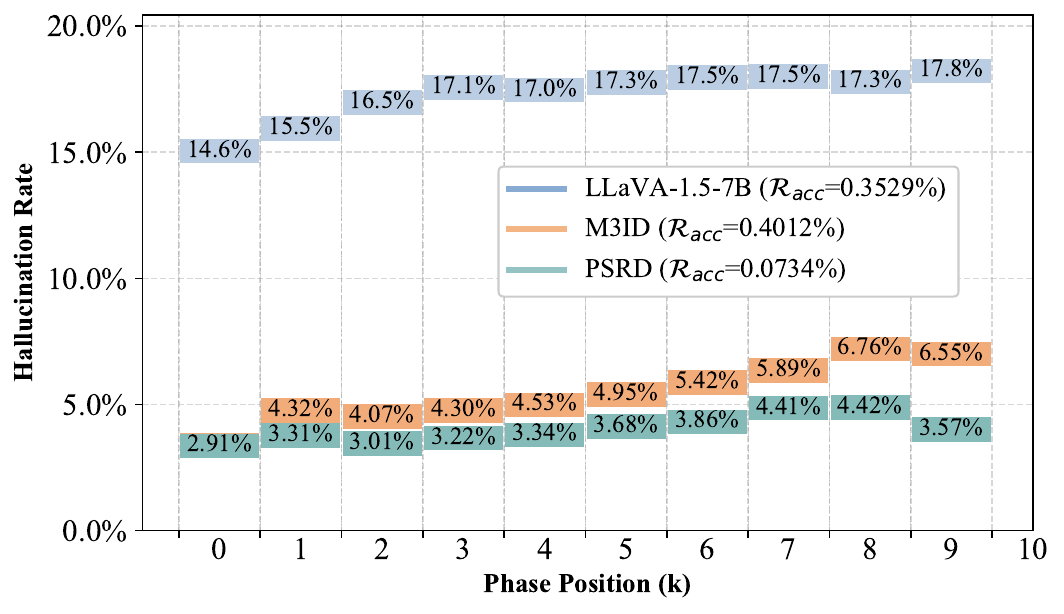}
\caption{
Quantitative results of phase-specific hallucination mitigation.
By intervening at critical phase junctions, PSRD suppresses hallucination propagation and achieves a lower $R_{\mathrm{acc}}$ than LLaVA-1.5-7B and M3ID.
At index 0, PSRD and M3ID exhibit identical hallucination rates.
} 
\label{fig:hallucination_change}
\end{figure}

For discriminative hallucination mitigation tasks, \textsc{PSRD} significantly improves the performance of LLaVA-1.5-7B by 13.9 and 5.4 percentage points in terms of F1-score on the AMBER and the full POPE benchmark in Table~\ref{tab:table2} and consistently surpasses other post-hoc baselines for nearly all evaluation metrics on both datasets. 
These results show that the proposed method can effectively mitigate hallucination across diverse settings, achieving performance comparable to methods trained with costly annotated data.

\subsection{Generalization of the Proposed Method}
\label{sec:generalization}
Beyond LLaVA-1.5-7B, we demonstrate the generalization capability of the proposed method across models of varying sizes and architectures, including InstructBLIP-7B, LLaVA-NeXT-7B, and LLaVA-1.5-13B.
As shown in Table~\ref{tab:other_model}, our method consistently yields the best results across all LVLMs, measured by CHAIR and Hal on the AMBER dataset.
Specifically, when applied to the InstructBLIP-7B model, our method reduces the CHAIR score by 27.9\% and the Hal by 5.9\% compared to Octopus. 
Crucially, while conceding only a minimal deficit in entity coverage (Cover) of 1.4\% against Octopus, this result underscores \textsc{PSRD}'s superior ability to preserve factual entity information throughout the hallucination mitigation process. 
For LLaVA-Next-7B, \textsc{PSRD} slightly reduces the Cover score while achieving substantial improvements on CHAIR, Hal, and Cog.
For larger-sized LVLMs, our method applied to LLaVA-1.5-13B substantially surpasses all baselines, achieving particularly notable hallucination reduction, as evaluated by CHAIR and Hal metrics.
These results demonstrate the advantage of generalizing the proposed method across a broader spectrum of LVLMs.

\subsection{Effectiveness of Hallucination Mitigation}
\label{sec:mitigation_effectiveness}
We quantify the effectiveness of phase-specific hallucination mitigation for \textsc{PSRD} using the hallucination rate defined in Section~\ref{sec:Dynamics}. 
We compare against M3ID, a dynamic sampling--based decoding method, and LLaVA1.5-7B.
As shown in Figure~\ref{fig:hallucination_change}, the proposed method reduces the average hallucination rate by $13.2\%$ across generation phases compared to LLaVA1.5-7B.
To quantify hallucination propagation during generation, we define the average phase-level hallucination accumulation rate ($\mathcal{R}_{acc}$), measuring the average increase in hallucination severity between consecutive phases.
The accumulation rate $\mathcal{R}_{acc}$ is defined as:

$$
\mathcal{R}_{acc} = \frac{1}{N-1} \sum_{i=1}^{N-1} (\text{CHAIR}_{i+1} - \text{CHAIR}_{i}),
$$
where $N$ is the total number of phases in the caption, and $\text{CHAIR}_i$ is the $\text{CHAIR}$ score of the phase at index $i$. 
A lower $\mathcal{R}_{acc}$ indicates a more stable and less propagating hallucination profile.
The $\mathcal{R}_{acc}$ achieved by LLaVA1.5-7B and M3ID is $0.35\%$ and $0.40\%$, respectively, which is approximately seven times higher than our method's $\mathcal{R}_{acc}$ ($0.07\%$).
This result underscores the efficacy of our stage-specific and progressively adaptive strategy in not only suppressing existing hallucinations but also actively preventing their propagation to subsequent generation steps.

\begin{table}[t]
    \centering
    \small
    \caption{Experimental results on the AMBER HalDet and MHal-detect datasets. We report the results including Precision (P), Recall (R), Accuracy (ACC), and F1-score (F1).}
    \resizebox{\linewidth}{!}{
    \begin{tabular}{lcccccccc}
        \toprule
        \multirow{2}{*}{\textbf{Model}} & \multicolumn{4}{c}{\textbf{AMBER HalDet}} & \multicolumn{4}{c}{\textbf{MHal-detect}} \\
        \cmidrule(lr){2-5} \cmidrule(lr){6-9}
        & ACC & R & P & F1 & ACC & R & P & F1 \\
        \midrule
        FG-CLIP  & 53.1 & \textbf{93.6}& 48.8 & 64.2  & 60.6 & \textbf{85.9} & 53.6 & 66.1 \\
        Open-CLIP & 75.0 & 89.2 & 80.7 &          \underline{84.7}& 68.9 & 77.1 & 80.4 & \underline{78.7}\\
        \textbf{Ours} & \textbf{80.5}& 88.6 & \textbf{88.9}& \textbf{88.7} &\textbf{72.0}& 76.5 &\textbf{87.8}&\textbf{81.7} \\
        \bottomrule
    \end{tabular}
    }
    \label{tab:comparison_results}
\end{table}

\begin{table}[t]
    \centering
    \small
    \caption{Ablation study on the AMBER and MHal-Detect datasets. "w/o" indicates the removal of a specific component.}

    \resizebox{\linewidth}{!}{

    \begin{tabular}{lcccccccc}
        \toprule
        \multirow{2}{*}{\textbf{Method}} & \multicolumn{4}{c}{\textbf{AMBER}} & \multicolumn{4}{c}{\textbf{MHal-detect}} \\
         \cmidrule(lr){2-5} \cmidrule(lr){6-9}
         &CHAIR $\downarrow$ & Cover $\uparrow$ & Hal $\downarrow$ &  Cog $\downarrow$  & P $\uparrow$ & F1 $\uparrow$\\
        \midrule
        Ours & \textbf{3.9}&\textbf{ 48.2} & 20.1 & 2.0  & \textbf{87.8} & \textbf{81.7} \\
        w/o $p_{con}$ & 4.7 & 46.0 & \textbf{20.0} & \textbf{1.8 }&  83.8 & 80.6 \\
        w/o $\mathcal{L}_{\text{DA}}$ & 4.9 & 46.2 & 20.9 & 1.9 &  73.9 & 76.5 \\
        w/o $\mathcal{L}_{\text{Margin}}$ & 5.3 & 46.9 & 23.9 & 1.4 &  79.8 & 79.0 \\
        w/o $\mathcal{L}_{\text{HC}}$ & 5.0 & 47.7 & 24.7 & 2.5 & 83.3 & 80.1 \\
        w/o VCD & 3.5 & 41.7 & 14.7 & 1.3 & - & - \\

        \bottomrule
    \end{tabular}
    }

    \label{tab:ablation_study}
\end{table}

\begin{figure}[t]
\includegraphics[width=0.95\linewidth]{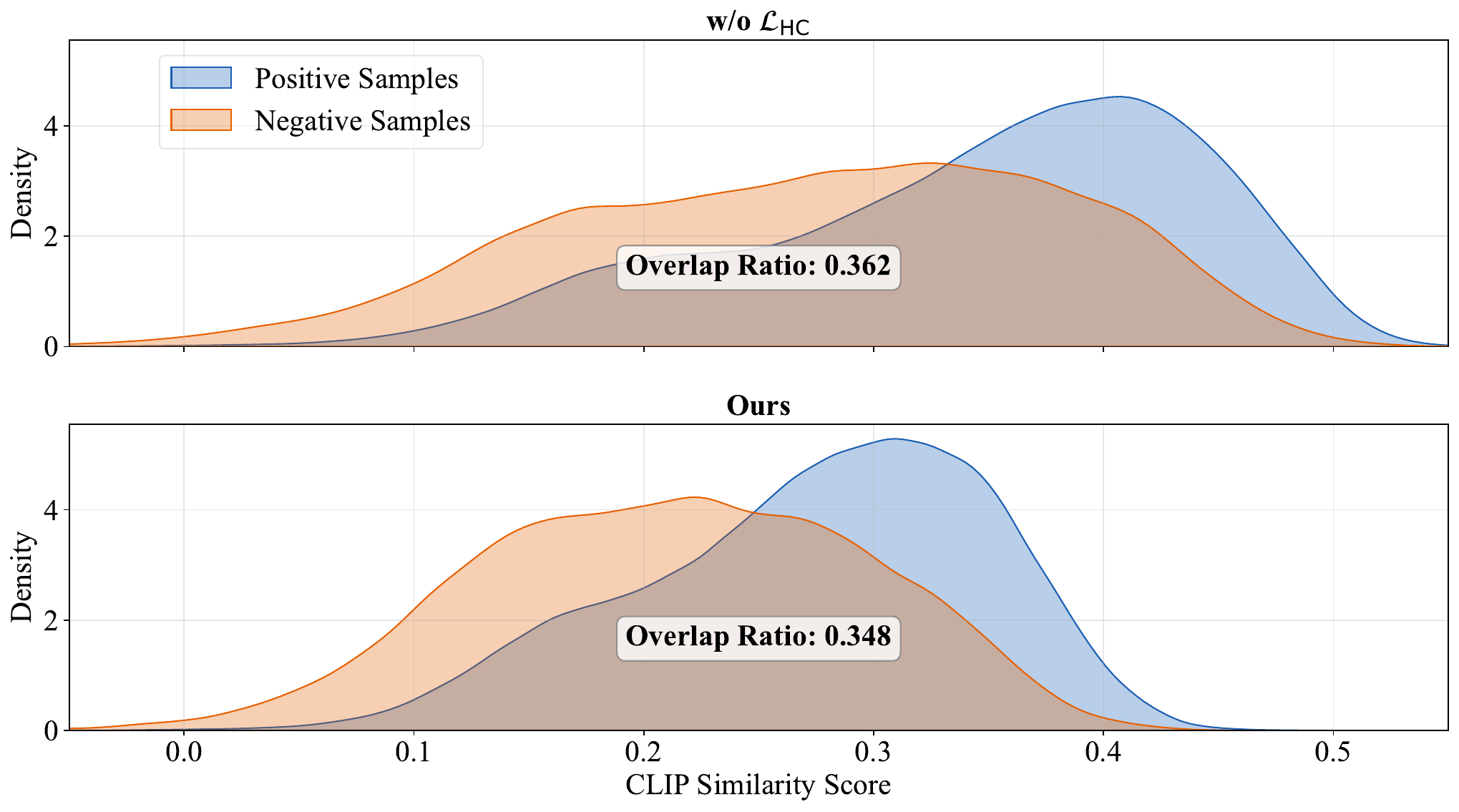}
\caption{
   Distribution of reward model output scores under different training settings. We visualize the density of CLIP similarity scores for positive and negative samples when training with and without the $\mathcal{L}_{\text{HC}}$ Loss. Incorporating $\mathcal{L}_{\text{HC}}$ leads to better separation between the two distributions, reflected by a reduced overlap ratio, indicating improved discriminability of the reward model.
} 
\label{fig:other}
\end{figure}

\subsection{Analysis for Reward Model}
\paragraph{\textbf{Hallucination Classification for Reward Model}}
The lightweight reward model ($\mathcal{R}$) is evaluated on two phase-level hallucination classification benchmarks, AMBER HalDet and MHal-Detect.
As shown in Table~\ref{tab:comparison_results}, $\mathcal{R}$ outperforms Open-CLIP~\cite{radford2021learning}, the base reward model, and FG-CLIP~\cite{xie2025fg}, which targets fine-grained visual--textual alignment.
Compared to Open-CLIP, $\mathcal{R}$ achieves gains of up to $8.9\%$ in accuracy and $4.3\%$ in F1.

Although $\mathcal{R}$ adopts CLIP as its architectural backbone, its training objective differs substantially from generic image--text alignment.
By incorporating uncertainty-weighted supervision distilled from the LVLM’s self-evaluation, $\mathcal{R}$ is optimized to capture distinctions between hallucinated and visually grounded descriptions, rather than relying solely on semantic similarity.
As a result, the learned reward signal reflects hallucination-related cues that are not explicitly encoded in standard CLIP representations.
This property allows $\mathcal{R}$ to serve as a phase-wise hallucination detector, providing stable and interpretable reward signals for targeted intervention during decoding.
The reliability of the underlying self-evaluation signals is further analyzed in Appendix~\ref{app:pseudo_label_reliability}.
\begin{figure}[t]
\centering
\includegraphics[width=0.95\linewidth]{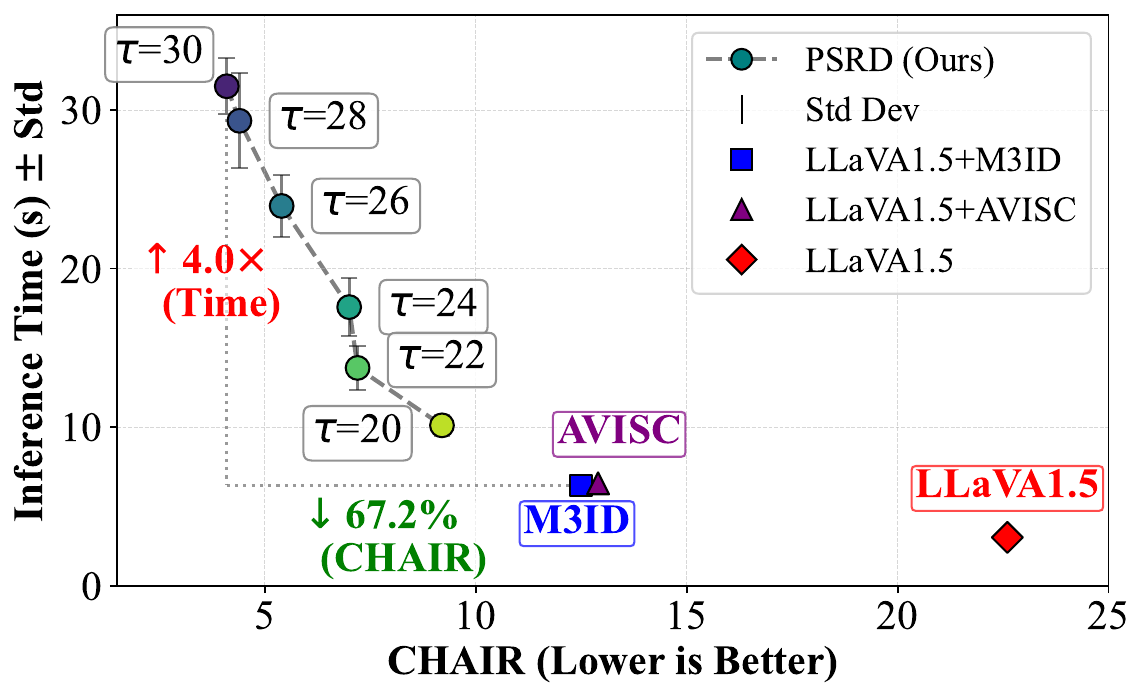}
\caption{
    Trade-off between hallucination mitigation effectiveness and efficiency of the proposed $\text{\textsc{PSRD}}$ on the \textit{Object HalBench}. 
    We vary the threshold $\tau$ in Sec.~\ref{sec:mitigation} and evaluate the performance of hallucination degree (CHAIR score, lower is better) and the average time consumed per question (in seconds, lower is better). 
    The compared methods are $\text{LLaVA1.5-7B}$~\cite{llava1.5}, $\text{M3ID}$~\cite{favero2024multi} and AVISC~\cite{woo2024don}.
} 
\label{fig:threshold}
\end{figure}

\paragraph{\textbf{Component Analysis of Reward Model.}}
We evaluate the contribution of the individual learning objectives and the uncertainty weighting mechanism through a detailed ablation study. 
The notation "w/o $p_{\text{con}}$" denotes the removal of the uncertainty weights ($p_{I,k}^+$, $p_{I,k}^-$ and $p_{I,k'}^-$) in the all loss functions.
The result is summarized in Table~\ref{tab:ablation_study} across the hallucination mitigation task (AMBER) and the hallucination classification benchmark (MHal-detect).

Integrating all components, the proposed method achieves the best overall performance with a CHAIR score of $3.9$ for AMBER and F1 score of $81.7$ on MHal-detect.
We observe that the removal of the uncertainty weighting mechanism, which is critical for suppressing noise from unreliable pseudo-labels, causes a clear degradation in both tasks (CHAIR increases by $0.7$ for AMBER and F1 drops by $1.1$ for MHal-detect). 
Eliminating $\mathcal{L}_{\text{DA}}$, which serves as the foundational objective, results in the most severe drop in hallucination detection, with F1 score plummeting by $5.2$. 
The removal of $\mathcal{L}_{\text{Margin}}$, designed to enforce robust feature separation, causes the most significant degradation in hallucination mitigation, evidenced by the CHAIR score increasing by $1.3$. 
Excluding $\mathcal{L}_{\text{HC}}$, which enforces intra-class compactness for hallucinated phases, leads to a notable increase in the Hal score to $24.7$, indicating reduced consistency in hallucination modeling.
Removing VCD in targeted intervention yields lower CHAIR and Hal scores but substantially degrades entity coverage, with Cover dropping from $48.2$ to $41.7$.

These results validate that our uncertainty-guided multi-objective design is essential for effective reward model training by capturing the fine-grained semantic differences required to distinguish hallucination from non-hallucination text. Further analyses of hyperparameter sensitivity and the rationale for the default loss weights are provided in Appendix~\ref{app:additional_experiments}, including the magnitude-balancing visualization in Figure~\ref{fig:loss_balancing_appendix}.
\begin{table}[t]
\centering
\caption{Ablation of hyperparameters in Efficient Reward-guided Targeted Intervention. We report CHAIR (lower is better) and Cover (higher is better) in the \textit{AMBER}~\cite{amber}  generative benchmark under different $(k,\delta,\alpha_{\max})$ settings. Results are grouped by the acceptance threshold $\tau$, and within $\tau{=}30$ further organized by $k$ for readability.}
\small
\setlength{\tabcolsep}{12pt}
\begin{tabular}{c c c c c}
\toprule
$k$ & $\delta$ & $\alpha_{\max}$  & CHAIR $\downarrow$ & Cover $\uparrow$ \\
\midrule

\multicolumn{5}{l}{\emph{ $k=1$ (vary $\alpha_{\max}$)}} \\
\cmidrule(lr){1-5}
1 & 0.5 & 2 & 8.5 & 53.9 \\
1 & 0.5 & 3 & 8.3 & 52.0 \\
1 & 0.5 & 4 & 8.4 & 53.5 \\
\midrule

\multicolumn{5}{l}{\emph{ $k=3$ (vary $\alpha_{\max}$)}} \\
\cmidrule(lr){1-5}
3 & 0.5 & 2 & 6.8 & 47.8 \\
3 & 0.5 & 3 & 6.7 & 47.1 \\
3 & 0.5 & 4 & 7.3 & 45.8 \\
\midrule

\multicolumn{5}{l}{\emph{ $k=5$ (vary $\delta$ and $\alpha_{\max}$)}} \\
\cmidrule(lr){1-5}
5 & 0.2 & 3 & 4.0 & 48.4 \\
5 & 0.5 & 3 & 3.9 & 48.2 \\
5 & 1.0 & 3 & 4.1 & 48.0 \\
\cmidrule(lr){2-5}
5 & 0.5 & 3 & 3.9 & 48.2 \\
5 & 0.5 & 4 & 3.9 & 45.9 \\

\bottomrule
\end{tabular}
\label{tab:erti_ablation}
\end{table}
\paragraph{\textbf{Analysis for Sample Discriminability of Reward Model.}}
We conduct an auxiliary analysis to evaluate the ability of reward model to distinguish between hallucinated (positive) and non-hallucinated (negative) samples. 
This is achieved by computing the hallucination score in Eq~\ref{eq:SCORE}, for both positive and negative phase-image pairs within the Amber HalDet dataset across different training settings.
Specifically, we plot the probability density functions of the hallucination scores for the positive and negative samples across the entire dataset. A highly effective reward model is characterized by a strong separation between the two distributions, which is quantitatively represented by a smaller area of overlap.

Based on results of Table~\ref{tab:ablation_study}, $\mathcal{L}_{HC}$ leads to the most substantial performance degradation upon removal (e.g., Hal increases from 20.1 to 24.7), indicating its critical role in hallucination suppression. We therefore focus our distributional analysis on the representative loss component.
As illustrated in Figure~\ref{fig:other}, we compare the impact of including the $\mathcal{L}_{HC}$ loss term on the reward model's performance. 
We observe that the addition of $\mathcal{L}_{HC}$ effectively \textbf{compresses the internal score variance} for both positive and negative samples, forcing the scores into a \textbf{significantly narrower numerical range} (indicated by a smaller span on the x-axis).

This reduction in score variability enhances the separation between positive and negative distributions, reducing their overlap area to 0.348.
This finding demonstrates that $\mathcal{L}_{HC}$ achieves superior hallucination detection by \textbf{encouraging greater intra-class compactness} (making scores within positive/negative groups closer) while simultaneously increasing the inter-class separation (pushing the two distributions further apart).

\subsection{Analyses and Discussions}
\paragraph{\textbf{Trade-off between Efficiency and Effectiveness.}}
\label{sec:tradeoff_analysis}
As demonstrated in Sect.~\ref{sec:threshold}, adjusting the threshold $\tau$ shifts the boundary where $\text{\textsc{PSRD}}$ determines whether a phase is a hallucination.
In this sense, the threshold functions as a reward-sensitivity control parameter that modulates the strength of the model’s response to self-evaluated reward feedback during inference.
We further analyze the inherent trade-off between effectiveness and efficiency under various thresholds using the Amber generation benchmark. 
We quantify the effectiveness of hallucination mitigation using the CHAIR score and measure the efficiency using the average inference time per question (seconds).
Specifically, we vary the threshold $\tau$ from $30\%$ to $20\%$ for the proposed $\text{\textsc{PSRD}}$. 
We also compare its performance against the base LLaVA1.5-7B and dynamic decoding methods (M3ID and AVISC). 

As shown in Figure~\ref{fig:threshold}, we observe that as the threshold $\tau$ decreases, which corresponds to a progressively more relaxed criterion for hallucination detection, the time consumed by $\text{\textsc{PSRD}}$ decreases, yet the hallucination severity increases, indicated by a higher $\text{CHAIR}$ score.
At $\tau=30\%$, the proposed $\text{\textsc{PSRD}}$ achieves a $67.2\%$ $\text{CHAIR}$ reduction over $\text{M3ID}$, with a $4.0\times$ inference-time increase.
This demonstrates a \textbf{highly controllable and flexible trade-off} between efficiency and effectiveness in PSRD.

\paragraph{\textbf{Hyperparameter Analysis for Target Intervention.}}
We conducted a systematic hyperparameter analysis to evaluate the sensitivity of the targeted intervention mechanism. This analysis focuses on the top-$k$ candidate decoding branches, the maximum continuous intervention strength $\alpha_{\max}$, and the probe step $\delta$ utilized in the secant-style update, all governed by a pre-defined reward acceptance threshold of $\tau = 30$.
The results of these ablations are summarized in Table~\ref{tab:erti_ablation}, evaluated using two primary metrics: CHAIR (measuring hallucination rate; lower is better) and Cover (measuring entity coverage and linguistic richness; higher is better).

The choice of $k$ controls the breadth of low-cost scouting and bounds the number of reward evaluations; $k{=}5$ provides sufficient branch exploration while keeping inference overhead modest.
The probe step $\delta$ determines the finite-difference estimate used by the secant update; we set $\delta{=}0.5$ as a stable, moderate step that avoids excessively small perturbations (which can be noisy under stochastic decoding) while remaining local enough for the update to be meaningful.
Finally, $\alpha_{\max}$ caps the maximum intervention strength to prevent overly aggressive contrastive penalties and to provide a predictable worst-case compute budget.

We set $\eta \approx 1.1$ due to the local concavity of the reward landscape, where marginal gains diminish quickly.
Strict linear extrapolation would therefore risk underestimating the required adjustment, potentially failing to meet the acceptance threshold.
The $10\%$ over-relaxation buffer accounts for this non-linearity, increasing the chance of single-step convergence. This slight over-intervention is preferable, as the cost of additional inference passes outweighs the minor effect of a marginally higher penalty weight.

\paragraph{\textbf{Analysis of Generated Content Quality.}}
\label{sec:content_quality}

Beyond the core hallucination mitigation performance, we conduct a further evaluation on the quality of the generated content produced by the proposed \textsc{PSRD}.
Specifically, we compare the \textbf{fluency, grammatical consistency, and overall readability} of the final generated paragraphs against the dynamic decoding method, $\text{M3ID}$~\cite{favero2024multi}. 
To ensure an objective and scalable comparison, we employ a human-like evaluation protocol using the powerful $\text{ChatGPT-4o-mini}$~\cite{gpt4} on 500 randomly selected cases from the Amber generative benchmark. 
The detailed instruction prompt provided to the LLM judge is specified in Supplementary Material.

The results show that the proposed \textsc{PSRD} is preferred over M3ID in \textbf{$48.5\%$} of cases, with M3ID being preferred in $37.5\%$ and the remaining $14.0\%$ rated as equally good. 
This finding confirms that $\text{\textsc{PSRD}}$ not only achieves superior hallucination mitigation but also maintains competitive generation quality.

\section{Related Work}
\label{sec:related_work}

\subsection{Multimodal Hallucination Mitigation}

LVLMs have demonstrated strong performance in visual perception~\cite{ma2024vision, zhang2025evaluating, zhang2025moma}, understanding~\cite{huang2026sat,wei2025first,zuo2025inimagetrans} and reasoning~\cite{lu2022learn,xu2026beyond, ma2026beyond,zhu2026decoupling,li2026dynamics}. 
However, they still suffer from severe vision hallucination~\cite{jiang2024hallucination,zhang2026instruction,leng2024mitigating}, which limits their application in real-world scenarios.
Existing efforts to mitigate multimodal hallucinations generally follow two primary trajectories: training-based alignment~\cite{fu2025chip,jiang2024hallucination,fu2025mitigating} and inference-time post-hoc correction~\cite{leng2024mitigating,lee2023volcano,favero2024multi}.

\paragraph{\textbf{Training-based Alignment.}}
A line of research focuses on fine-tuning LVLMs using preference-based datasets to align model outputs with visual ground truth. 
Early methods rely on extensive human labeling or knowledge distillation from superior proprietary models~\cite{fu2025chip,jiang2024hallucination}. 
For instance, HSA-DPO~\cite{xiao2025detecting} and HDPO~\cite{fu2025mitigating} construct high-quality preference pairs to isolate specific failure modes.
RLAIF-V~\cite{Yu_2025_CVPR} and EOS~\cite{yue2024less} utilize open-source models for preference feedback, while self-improvement frameworks~\cite{deng2024enhancing, tan2025beyond} iteratively train LVLMs on self-generated data. 
Despite their efficacy, these approaches incur prohibitive computational overhead and are constrained by the quality of the training data.

\paragraph{\textbf{Post-hoc Mitigation Strategies.}}
Post-hoc methods aim to suppress hallucinations during or after the inference process, which can be further categorized into two paradigms:
(1) \textbf{Generate-then-revise methods} typically employ a multi-step pipeline to correct the initial response. Volcano~\cite{lee2023volcano} utilizes natural language feedback to self-correct initial outputs via predefined prompts. Woodpecker~\cite{yin2024woodpecker} further expands this into a comprehensive pipeline involving concept extraction, visual verification, and claim-level correction driven by external LLMs. While effective, these methods often suffer from high inference latency due to multiple decoding passes.
(2) \textbf{Contrasting decoding methods} recalibrate the output distribution during a single forward pass. A prominent research direction involves incorporating contrastive signals~\cite{leng2024mitigating,wang2024mitigating} or penalizing specific attention patterns, such as blind summary tokens in OPERA~\cite{huang2024opera,woo2024don}. Other techniques involve manipulating visual features by zeroing out hallucinatory image tokens~\cite{che2025hallucinatoryimagetokenstrainingfree, an2025mitigating} or leveraging external models to provide guidance signals~\cite{park2025convis, wan2025only}. Recently, adaptive decoding has emerged, where strategies are switched based on context~\cite{chen2025mixture}, hallucination types~\cite{suo2025octopus}, or internal information metrics like multi-modal mutual information~\cite{favero2024multi}. 

However, these approaches often overlook the dynamic nature of hallucination emergence during decoding. In contrast, \textsc{PSRD} introduces a phase-wise self-reward mechanism to facilitate adaptive, targeted intervention at the critical onsets of semantic segments—precisely where the risk of hallucination is most pronounced.

\subsection{Reward-guided Controlled Generation}
Reward-guided decoding~\cite{arora2206director,yang2021fudge,schulman2017proximal,rame2024warm} has emerged as a powerful paradigm for controlled generation in LLMs. 
Early methods such as FUDGE~\cite{yang2021fudge} and DIRECTOR~\cite{arora2206director} utilize auxiliary prefix scorers to steer the decoding process toward specific constraints, while reinforcement learning frameworks like PPO~\cite{schulman2017proximal} align model outputs with human preferences through reward-based feedback. Recently, MRGD~\cite{manas2025controlling} extended this concept to LVLMs to mitigate hallucinations during inference. However, MRGD relies on extensive external annotation data to train its dual reward models, which incurs significant labeling costs and limits its generalizability.

\section{Discussion}
\label{supp:disc}

More broadly, our findings advocate for a process-oriented perspective on hallucinating mitigation via self-reward modeling. A fundamental insight is that discrimination is structurally more tractable than generation: while faithful generation necessitates long-horizon grounded synthesis, identifying unreliable intermediate outputs often benefits from stronger priors and denser supervision. This asymmetry suggests that training robust discriminators to guide the generative process is a more effective pathway than solely relying on scaling the generator itself.

From this vantage point, self-rewarding serves not only as an inference-time hallucination mitigation strategy but also as a lightweight paradigm for model self-evolution. Instead of the computationally expensive continual retraining of the base LVLM, the system's overall performance can be enhanced by developing specialized reward models through structured supervision, such as confidence calibration, process-level consistency, and grounding-oriented objectives. These reward models provide fine-grained guidance during inference, allowing the frozen base model to exhibit superior capabilities without any parameter updates.

We believe this paradigm establishes a scalable and promising trajectory for building more reliable, controllable, and self-improving LVLMs.

\section{Conclusion}
In this paper, we introduce a new self-rewarding framework, enabling dynamic hallucination mitigation at inference time without external supervision. 
On the empirical side, we reveal that visual hallucination exhibits phase-wise dynamic patterns, peaking at the onset of each semantic phase. 
Inspired by these insights, we propose \textbf{\textsc{PSRD}} for online hallucination correction guided by phase-wise self-reward signals.
Specifically, we first distill the guidance signal into a lightweight reward model to reduce the cost of online repeated self-evaluation during decoding.
The reward model subsequently provides on-the-fly guidance for targeted intervention during the decoding process, enabling precise hallucination suppression.
The proposed \textsc{PSRD} significantly reduces hallucinations in LLaVA-1.5-7B by 50.0\% and consistently outperforms existing post-hoc methods across five benchmarks for four LVLMs.
Further analysis confirms that \textsc{PSRD} effectively mitigates hallucination propagation and achieves a highly controllable trade-off between strong performance and inference efficiency.

\bibliographystyle{ACM-Reference-Format}
\bibliography{sample-base}

\clearpage
\appendix

Our supplementary materials are summarized as follows:
\begin{itemize}[itemsep=5pt,topsep=1.5pt,leftmargin=10pt]
    \item Appendix~\ref{supp:metric_hall}: Details for Hallucination Rate in Section~\ref{sec:Dynamics}.
    \item Appendix~\ref{supp:generation}: Details for Uncertainty-Guided Data Generation without external annotations or supervision.
    \item Appendix~\ref{supp:method}: Method implementation details including reward model training  and  dynamic hallucination mitigation.
    \item Appendix~\ref{app:additional_experiments}: Additional Analyses of the proposed PSRD and the reward model.
    \item Appendix~\ref{supp:exp}: Baselines and detailed experiment settings in Section~\ref{sec:experiments}.
\end{itemize}

\section{Details for Hallucination Rate in Section~\ref{sec:Dynamics}}
\label{supp:metric_hall}
In this section, we provide a detailed description of the phase-level hallucination rate and the word-level hallucination rate used to quantify hallucination severity in Section~\ref{sec:Dynamics}.

Let $H_{i,k,j}$ be the binary word hallucination indicator function for a given sample $i$ at phase position $k$ and normalized word position $j$:
\begin{equation}
H_{i,k,j} = 
\begin{cases}
1 & \text{if word at position } (k,j) \text{ is hallucinated} \\
0 & \text{otherwise}
\end{cases}
\end{equation}

The word-level hallucination rate $R_{\text{word}}(k,j)$ at position $(k,j)$ is defined as the empirical probability of a word being hallucinatory across all samples:
\begin{equation}
R_{\text{word}}(k,j) = \frac{1}{m} \sum_{i=1}^{m} H_{i,k,j}
\end{equation}
where $k$ is the phase position index (phase index), $j$ is the normalized word position within phase $k$, where $j \in [0,1]$, and $m$ is the total number of samples evaluated.

The phase-level hallucination rate at position $k$ represents the probability that a phase at position $k$ contains at least one hallucinated word, calculated over $m$ samples. 
Let $S_{i,k}$ be the phase hallucination indicator function for sample $i$:
\begin{equation}
S_{i,k} = 
\begin{cases} 
1 & \text{if phase } k \text{ is hallucinated} \\
0 & \text{otherwise}
\end{cases}
\end{equation}

The phase-level hallucination rate $R_{\text{sent}}(k)$ at position $k$ is then:

\begin{equation}
R_{\text{sent}}(k) = \frac{1}{m} \sum_{i=1}^{m} S_{i,k}
\end{equation}

\section{Details for Uncertainty-Guided Data Generation}
\label{supp:generation}
\subsection{Generating Hallucinated Image Captions}

We use the COCO2014~\cite{lin2014microsoft} training set as the image source for caption generation.  
To systematically elicit hallucination behavior in LVLMs, we adopt two complementary strategies:
\paragraph{Image Corruption}
We introduce vision hallucinations by adding zero-mean Gaussian noise to input images, with standard deviation $\sigma \in [0.2, 0.6]$, degrading the visual signal while preserving overall semantics.

\paragraph{Instruction Corruption}
To elicit controlled hallucinations through instruction corruption, we manipulate textual prompts to first anchor responses in observable image content, then extend them with logically coherent yet non-existent elements. This structured prompting encourages the model to blend factual and fabricated details in a fluent and contextually plausible manner. Illustrative prompts are shown as follows:
\begin{flushleft}
\vspace{+0.3cm}
\hrule height 1pt 
\vspace{+0.1cm}
\noindent \textbf{Prompt: Hallucination elicitation with instruction corruption} 

\begin{quote} 
\small
 Describe the image in detail, first focusing on the key elements that are clearly present. Then, enrich your description by suggesting plausible additions—objects, characters, or environmental details that could logically exist in or around the scene. While creatively expanding the narrative, ensure that all of your description remains grounded in observable features of the image. Generate responses in a concise style, avoiding excessive embellishment or complexity. \\

\end{quote}

 \hrule height 1pt 

\end{flushleft}

\subsection{Details for Phase-wise Uncertainty Signals Construction}
To construct training data for the reward model, we leverage the LVLM itself for phase-level annotation. 
We leverage punctuation marks (e.g., commas and periods) and conjunctions as candidate segmentation cues, and further employ spaCy as a syntactic parser to verify whether a split corresponds to a genuine semantic boundary. 

The LVLM is prompted to assess its alignment with the input image. 
The verification process includes: (1) object-level checks to detect common visual hallucinations, and (2) full-phrase assessment for broader inconsistencies. This self-annotation pipeline yields reliable (image, phrase, label) triplets without requiring human supervision. The automated labeling prompt template is provided below:

\begin{flushleft}

\vspace{+0.3cm}
\hrule height 1pt 
\vspace{+0.1cm}
\noindent \textbf{Prompt: Vision Hallucination Self-evaluation} 

\begin{quote} %
\small
Given an image and a phrase: '\{phrase\}', carefully verify if the phrase accurately describes the image by checking: 1) whether all specified objects (\{objects list\}) are present, 2) their quantities match if mentioned, 3) their spatial relationships if described, and 4) any attributes mentioned (colors, sizes, etc.). Respond strictly with 'Yes' only if every aspect of the phrase perfectly matches the image, otherwise respond 'No'. \\

\end{quote}

\vspace{+0.3cm}
\hrule height 1pt 
\vspace{+0.1cm}

\end{flushleft}

\section{Method Details}
\label{supp:method}

\subsection{Details of Reward Model Training}
For the three loss components in Equation~\ref{eq:loss_total}, we set the weights as $\lambda_1 = 1.0$, $\lambda_2 = 2.4$, and $\lambda_3 = 0.1$. The margin $\delta$ for the $L_{Margin}$ is set to 0.3. 
These weights are chosen to ensure that the Discriminative Alignment Loss dominates training, effectively separating hallucinated and non-hallucinated samples, while the regularization term helps maintain representation consistency across hallucinated content within the same image.
The reward model is fine-tuned for 5 epochs using the SGD\cite{amari1993backpropagation} optimizer with a constant learning rate of 1e-4 and a batch size of 64, distributed across 8 $\times$ RTX 4090 GPUs, completing in 7.5 hours. The training dataset comprises approximately 400k non-hallucinated examples and 40k hallucinated examples.

\subsection{Selection of Intervention Primitive in Reward-guided Targeted Intervention.}

While existing contrastive decoding approaches such as M3ID~\cite{favero2024multi} and AVISC~\cite{woo2024don} instantiate decoding-time interventions through complex attention-level manipulation mechanisms, which impose strong model-specific inductive biases on the generation process, such design choices may entangle the effects of the intervention mechanism with those of the underlying bias assumptions, thereby obscuring the generalization ability of the overall framework and reducing interpretability.
In contrast, our method employs only VCD~\cite{leng2024mitigating} as a minimal yet effective contrastive decoding primitive, allowing the performance gains to be attributed directly to PSRD.

\subsection{Reward-guided Targeted Intervention Algorithm.}
\label{sec:InterAlgorithm}
Algorithm~\ref{alg:search} presents the reward-guided targeted intervention procedure used in PSRD.
Given a lightweight reward model $\mathcal{R}$ and an acceptance threshold $\tau$, the goal is to identify a satisficing intervention configuration $(k,\alpha)$ such that $\mathcal{R}(k,\alpha)>\tau$ with a small number of reward evaluations.

The algorithm follows a two-stage \emph{Scout-and-Project} strategy.
In the first stage, we perform low-cost greedy scouting over the top-$K$ candidate decoding branches with zero intervention strength ($\alpha=0$).
This design is motivated by our empirical finding that hallucinations are more likely to emerge near the initial token of each phase, where the semantic trajectory is still being established.
PSRD therefore first probes a small set of candidate starting branches, evaluates their initial rewards $\mathcal{R}(k,0)$, and ranks them accordingly.
If any candidate already exceeds the threshold, the search terminates immediately.

In the second stage, the ranked candidates are examined one by one through bounded local refinement over the intervention strength $\alpha$.
For each candidate branch, PSRD starts from a small probe step and estimates a local slope from the observed reward change, which is then used to project the next value of $\alpha$.
This secant-style update is used as a local search mechanism for efficient refinement, rather than as a globally convergent optimization procedure.
Accordingly, PSRD does not assume a globally smooth or monotonic reward landscape in the discrete decoding space.
A relaxation factor is applied to slightly overshoot the estimated threshold-crossing point, which reduces repeated probing in practice.

To ensure stable refinement, PSRD explicitly checks the local reward trend during projection.
When the estimated slope becomes non-positive, the current refinement branch is terminated and the algorithm proceeds to the next ranked candidate.
This design is consistent with the role of the projection step as a bounded local heuristic: refinement is performed only when the observed reward change provides a meaningful basis for further projection.
The search is further bounded by a maximum intervention strength $\alpha_{\max}$ to avoid overly aggressive penalties and to provide a predictable worst-case computational budget.

Overall, the Scout-and-Project procedure serves as a practical and safeguarded local refinement strategy for early phase correction during decoding.

\begin{algorithm}[h]
   \caption{Scout-and-Project Threshold Search}
   \label{alg:search}
\begin{algorithmic}
   \STATE {\bfseries Input:} Reward Model $\mathcal{R}$, Threshold $\tau$, Candidates $K$, Probe Step $\delta$
   \STATE {\bfseries Output:} Optimal parameters $(k, \alpha)$
   
   \STATE \textcolor{gray}{// Stage 1: Greedy Scouting}
   \FOR{$k=0$ {\bfseries to} $K-1$}
       \STATE $s_k \leftarrow \mathcal{R}(k, 0)$
       \IF{$s_k > \tau$}
           \STATE \textbf{return} $(k, 0)$
       \ENDIF
   \ENDFOR
   
   \STATE $\mathcal{K}_{sorted} \leftarrow$ sort $k$ by $s_k$ in descending order
   
   \STATE \textcolor{gray}{// Stage 2: Bounded Local Projection}
   \FOR{each $k \in \mathcal{K}_{sorted}$}
       \STATE $s_{base} \leftarrow s_k$
       \STATE $\alpha_{curr} \leftarrow \delta$
       \STATE $s_{curr} \leftarrow \mathcal{R}(k, \alpha_{curr})$
       
       \WHILE{$s_{curr} \le \tau$ AND $\alpha_{curr} < \alpha_{max}$}
           \STATE $m \leftarrow (s_{curr} - s_{base}) / \alpha_{curr}$
           \IF{$m \le 0$}
               \STATE \textbf{break} \textcolor{gray}{// non-monotonic, try next $k$}
           \ENDIF
           \STATE $\Delta \alpha \leftarrow (\tau - s_{curr}) / m$
           \STATE $\alpha_{next} \leftarrow \alpha_{curr} + 1.1 \cdot \Delta \alpha$
           \STATE $s_{base} \leftarrow s_{curr}$
           \STATE $\alpha_{curr} \leftarrow \alpha_{next}$
           \STATE $s_{curr} \leftarrow \mathcal{R}(k, \alpha_{curr})$
       \ENDWHILE
       
       \IF{$s_{curr} > \tau$}
           \STATE \textbf{return} $(k, \alpha_{curr})$
       \ENDIF
   \ENDFOR
   \STATE $k^* \leftarrow \operatorname*{arg\,max}_k \{s_k\}$

   \STATE \textbf{return} $(k^*, 0)$ 
\end{algorithmic}
\end{algorithm}

\section{Additional Experimental Results}
\label{app:additional_experiments}
In this section, we provide additional analyses and control experiments that further clarify the behavior of PSRD, including local reward trends and bounded refinement behavior (\S\ref{app:monotonicity_fallback}), fluency under different intervention strengths (\S\ref{app:fluency_alpha}), hyperparameter sensitivity (\S\ref{app:hyperparameter_sensitivity}), the importance of early-phase intervention (\S\ref{app:early_vs_delayed}), reliability analysis of self-evaluation signals (\S\ref{app:pseudo_label_reliability}), phase-boundary definition via comparison with entropy-based segmentation (\S\ref{app:entropy_boundary}), an additional raw-CLIP baseline (\S\ref{app:clip_raw}), and a comparison between natural and induced hallucinations (\S\ref{app:artificial_natural}).

\subsection{Local Reward Trend and Bounded Refinement Behavior}
\label{app:monotonicity_fallback}

\begin{figure}[t]
  \centering
  \includegraphics[width=0.97\linewidth]{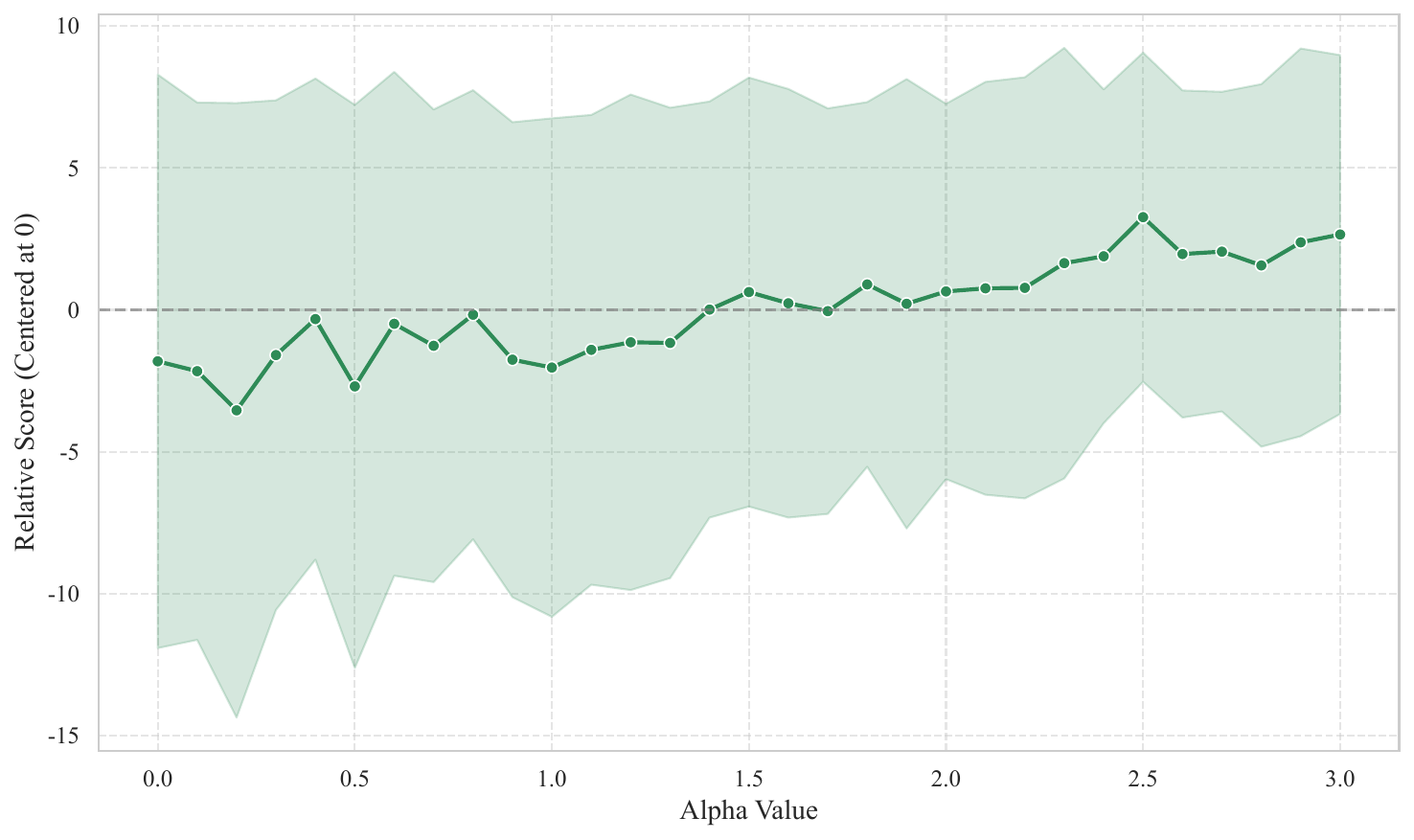}
  \caption{Reward score as a function of the intervention strength $\alpha$ within the bounded probing interval. PSRD does not require global monotonicity in the full decoding space; it only relies on a local empirical trend and uses a fallback branch when projection is unreliable.}
  \label{fig:alpha_reward_score_appendix}
\end{figure}

As discussed in the main text, PSRD does not rely on global monotonicity over the full discrete decoding space.
Instead, its intervention strategy is built on bounded local refinement, where the reward trend is only used within a small probing interval after candidate scouting.
To illustrate this behavior, Figure~\ref{fig:alpha_reward_score_appendix} visualizes the reward score under different intervention strengths $\alpha$.

Empirically, across 145 sentence groups, the reward score exhibits a clear local trend in the majority of cases, with 115 groups showing an increasing pattern within the probed interval.
This observation supports the use of local projection as an efficient refinement mechanism for a substantial portion of decoding branches.
At the same time, PSRD does not require this pattern to hold universally, since the projection step is applied only when the observed local reward change provides a useful basis for refinement.

The search procedure is always bounded within $[0,\alpha_{\max}]$.
When the local trend does not support further projection, PSRD terminates the current branch and proceeds to the next ranked candidate.
In this way, bounded refinement and branch switching work together as complementary components of the Scout-and-Project procedure, allowing PSRD to exploit favorable local reward structure when available while maintaining stable behavior across heterogeneous decoding trajectories.

\subsection{Fluency Impact Under Different Intervention Strengths}
\label{app:fluency_alpha}

\begin{figure}[t]
  \centering
  \includegraphics[width=0.97\linewidth]{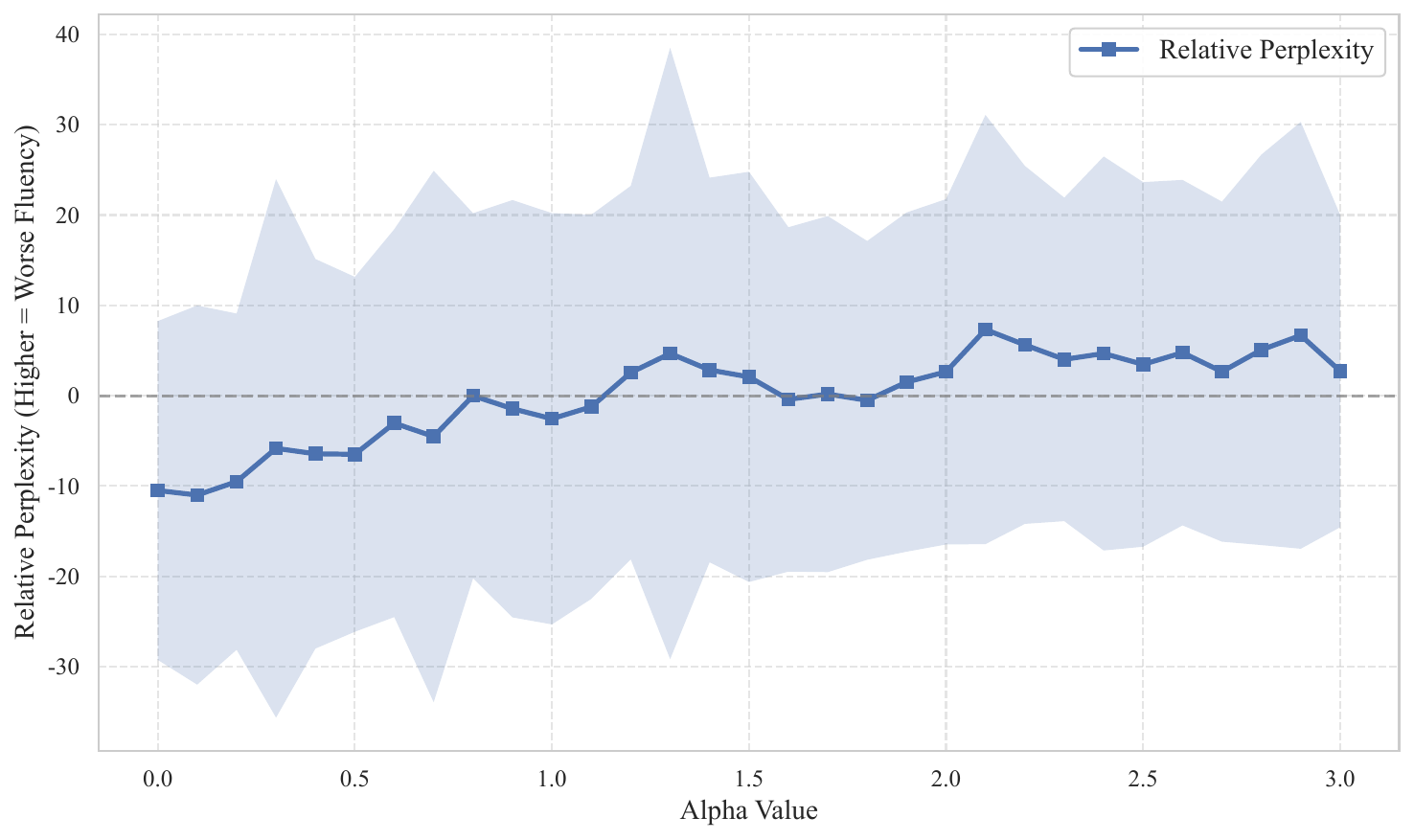}
  \caption{Relative fluency / perplexity change under different intervention strengths $\alpha$. The figure illustrates the reward--fluency trade-off during decoding.}
  \label{fig:alpha_fluency_appendix}
\end{figure}

We further analyze the fluency impact of different intervention strengths.
Figure~\ref{fig:alpha_fluency_appendix} shows the relative perplexity change under different $\alpha$.
Across the probed range, the overall fluctuation is limited, suggesting that the intervention strength has only a moderate effect on generation fluency.
While larger $\alpha$ is associated with a slight increase in perplexity in some regions, we do not observe a pronounced deterioration in fluency.
This result indicates that PSRD remains relatively stable under different intervention strengths, and that bounded search is sufficient to control intervention strength without sacrificing linguistic quality.

\subsection{Hyperparameter Sensitivity}
\label{app:hyperparameter_sensitivity}

We analyze the sensitivity of the reward-model training hyperparameters $(\lambda_1, \lambda_2, \lambda_3, \delta)$. These four hyperparameters are used only when training the reward model and are not introduced as test-time tuning variables during decoding. Throughout all experiments in the main paper and appendix, we fix a single default setting, $(1.0, 2.4, 0.1, 0.3)$, across all LVLMs and benchmarks.

To examine whether this choice is overly fragile, we train reward models with different hyperparameter configurations and then plug each trained reward model into the same PSRD framework with LLaVA-1.5-7B. The resulting system is evaluated on the AMBER generative benchmark~\cite{amber}, and the results are reported in Table~\ref{tab:hyperparameter_grid_appendix}. Overall, the default setting lies in a reasonably stable region rather than a narrow isolated optimum, suggesting that these hyperparameters mainly serve to balance the numerical scales of the three training losses instead of overfitting a specific benchmark. Across configurations with $\delta$ varying from 0.05 to 0.4, the resulting performance remains relatively stable, indicating that the method is not particularly sensitive to the precise choice of margin within this range. This further supports that the adopted default value $\delta=0.3$ is a robust choice rather than a narrowly tuned optimum.

\begin{table*}[t]
  \centering
  \caption{Sensitivity analysis over reward-model hyperparameters $(\lambda_1, \lambda_2, \lambda_3, \delta)$}
  \label{tab:hyperparameter_grid_appendix}
  \setlength{\tabcolsep}{6pt}
  \renewcommand{\arraystretch}{1.08}
  \begin{tabular}{c c cccc}
    \toprule
    ID & $(\lambda_1,\lambda_2,\lambda_3,\delta)$ & CHAIR $\downarrow$ & Cover $\uparrow$ & Hal $\downarrow$ & Cog $\downarrow$ \\
    \midrule
    1  & (1.0, 6.0, 0.4, 0.3)   & 4.0 & 48.1 & 18.8 & 1.9 \\
    2  & (1.0, 4.8, 0.2, 0.3)   & 4.8 & 48.4 & 18.8 & 1.9 \\
    3  & (1.0, 2.4, 0.4, 0.3)   & 4.7 & 46.8 & 20.0 & 1.7 \\
    4  & (2.0, 2.4, 0.2, 0.3)   & 4.1 & 48.1 & 20.0 & 1.6 \\
    5  & (1.0, 2.4, 0.1, 0.05) & 4.5 & 48.4 & 19.6 & 2.2 \\
    6  & (1.0, 2.4, 0.1, 0.4)  & 4.5 & 49.2 & 19.6 & 2.0 \\
    7  & (2.0, 2.4, 0.1, 0.3)  & 4.7 & 48.8 & 20.4 & 2.1 \\
    8  & (1.0, 4.8, 0.2, 0.2)   & 4.5 & 48.4 & 22.1 & 1.9 \\
    9  & (1.0, 4.8, 0.1, 0.3)  & 4.5 & 47.4 & 21.3 & 1.9 \\
    10 & (1.0, 2.4, 0.1, 0.5)  & 4.5 & 49.2 & 19.6 & 2.0 \\
    11 & (1.0, 2.4, 0.1, 0.1)  & 4.4 & 49.0 & 20.8 & 1.9 \\
    12 & (2.0, 6.0, 0.4, 0.3)   & 5.1 & 49.3 & 22.9 & 2.3 \\
    13 & (1.0, 2.4, 0.1, 0.2)  & 5.2 & 48.3 & 23.8 & 1.9 \\
    14 & (1.0, 2.4, 0.1, 0.3)  & 3.9 & 48.2 & 20.1 & 2.0 \\
    \bottomrule
  \end{tabular}
\end{table*}

\begin{figure*}[t]
  \centering
  \includegraphics[width=1.0\textwidth]{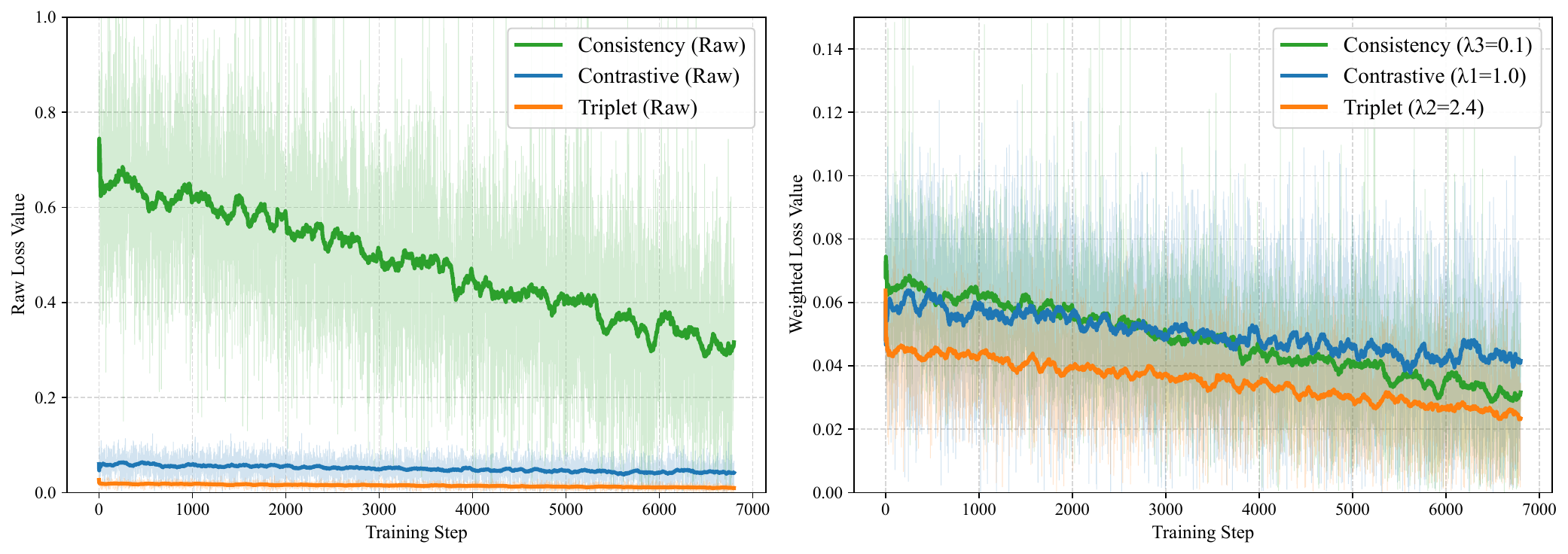}
  \caption{Justification of the default loss weights by magnitude balancing at the early training stage. 
  The three curves correspond to the Hallucination Consistency loss, the Discriminative Alignment loss, and the Margin Enforcement loss, respectively. 
  Before weighting, the consistency-related term dominates due to its larger numerical scale. 
  After applying the default weights $(\lambda_1, \lambda_2, \lambda_3)=(1.0, 2.4, 0.1)$, the magnitudes become more balanced, which helps stabilize multi-objective training.}
  \label{fig:loss_balancing_appendix}
\end{figure*}

To further clarify the choice of loss weights $(\lambda_1, \lambda_2, \lambda_3)$, we visualize the training dynamics of the three reward-model objectives at the early stage of training in Figure~\ref{fig:loss_balancing_appendix}. Before weighting, the consistency loss has a substantially larger numerical scale than the contrastive and triplet terms, and therefore dominates the optimization. After applying the default weights $(\lambda_1, \lambda_2, \lambda_3)=(1.0, 2.4, 0.1)$, the magnitudes of the three losses become more balanced, leading to comparable gradient contributions during training.

This result supports our design choice that the default weights are primarily selected for \emph{magnitude balancing} across objectives, rather than for benchmark-specific overfitting. In other words, the weighting is intended to prevent one numerically dominant term from overwhelming the others, so that the discriminative, margin-based, and consistency objectives can all contribute meaningfully to reward-model learning.

\subsection{Control Study: Early vs.\ Delayed Intervention}
\label{app:early_vs_delayed}

To verify that PSRD specifically benefits from intervention near phase boundaries, we compare the default early-phase intervention with a delayed-intervention control. In the delayed setting, the intervention starts from a random position between token 1 and the middle token of the phase, while the previously generated prefix is kept unchanged.

As shown in Table~\ref{tab:entropy_boundary_appendix}, delayed intervention is clearly less effective. This supports the hypothesis that phase boundaries are particularly vulnerable points, where the semantic trajectory is still being formed.

\subsection{Reliability of Self-Evaluation Signals}
\label{app:pseudo_label_reliability}

Since the reward model is trained with self-evaluated weak supervision, we further assess the reliability of the resulting pseudo-labels against an external reference. Specifically, we use LLaVA-1.5-7B as the hallucination evaluator and adopt the AMBER evaluation tool~\cite{amber} as the reference. On 3,407 samples, the pseudo-labels achieve an accuracy of 0.8739, a precision of 0.9012, a recall of 0.8944, and an F1 score of 0.8978.

\begin{figure}[t]
  \centering
  \includegraphics[width=0.99\linewidth]{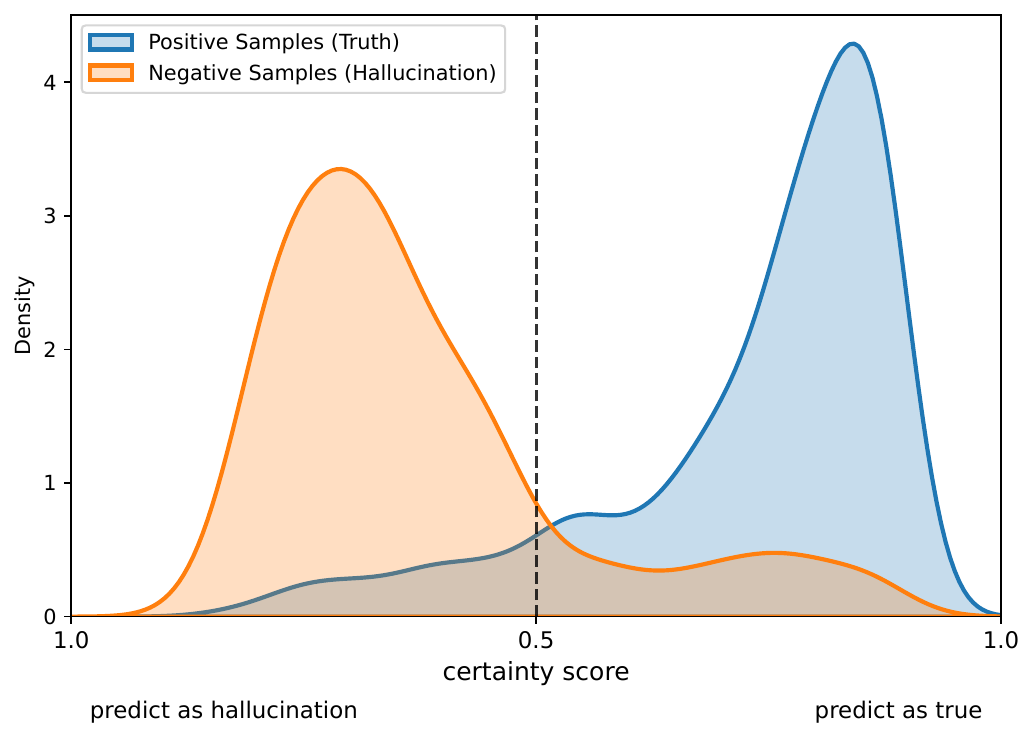}
  \caption{Distribution of pseudo-label confidence scores produced by the self-evaluation process. The overall separation between grounded and hallucinated samples supports the use of these signals as weak supervision.}
  \label{fig:pseudo_label_distribution_appendix}
\end{figure}

These results suggest that the self-evaluation signal is reasonably reliable as weak supervision. We further visualize the score distribution in Figure~\ref{fig:pseudo_label_distribution_appendix}.

\begin{table}[t]
  \centering
  \caption{Performance under adversarial / high-confidence hallucination prompts on AMBER generative~\cite{amber} with LLaVA-1.5-7B.}
  \label{tab:high_conf_hallucination_appendix}
    \renewcommand{\arraystretch}{1.12}
  \setlength{\tabcolsep}{5pt}
\resizebox{1.0\linewidth}{!}{
  \begin{tabular}{lcccc}
    \toprule
    Method & CHAIR $\downarrow$ & Cover $\uparrow$ & Hal $\downarrow$ & Cog $\downarrow$ \\
    \midrule
    \textbf{LLaVA-1.5-7B} & 9.0 & 51.6 & 42.5 & 3.7 \\
    + PSRD & 5.0 & 48.9 & 22.0 & 1.2 \\
    \bottomrule
  \end{tabular}}
\end{table}

\subsection{Comparison with Entropy-Based Phase Segmentation}
\label{app:entropy_boundary}

To examine whether a simple phase-boundary definition is sufficient, we additionally implement an entropy-based variant that determines phase transitions according to changes in token-level uncertainty, instead of using textual delimiters.

As shown in Table~\ref{tab:entropy_boundary_appendix}, the punctuation-based segmentation used in PSRD achieves performance comparable to this more complex entropy-based alternative on AMBER generative~\cite{amber}. This result suggests that the proposed phase partition strategy is already simple and effective, and that accurate intervention mainly depends on identifying vulnerable transition points rather than requiring a more elaborate boundary detector.

\begin{table}[t]
  \centering
  \caption{Performance under different phase-boundary definitions and intervention timings on AMBER generative.}
  \label{tab:entropy_boundary_appendix}
  \renewcommand{\arraystretch}{1.12}
  \setlength{\tabcolsep}{5pt}
  \resizebox{0.96\linewidth}{!}{
  \begin{tabular}{lcccc}
    \toprule
    Method & CHAIR $\downarrow$ & Cover $\uparrow$ & Hal $\downarrow$ & Cog $\downarrow$ \\
    \midrule
    \multicolumn{5}{l}{\textbf{LLaVA-1.5-7B w/ PSRD}} \\
    + punctuation boundary   & 3.9 & 48.2 & 20.1 & 2.0 \\
    + entropy-based boundary & 4.1 & 46.6 & 19.1 & 1.8 \\
    + delayed intervention & 6.1 & 47.5 & 33.6 & 3.3 \\

    \bottomrule
  \end{tabular}}
\end{table}

\subsection{Additional Baseline: Raw CLIP Similarity}
\label{app:clip_raw}

We further add a baseline that directly uses raw CLIP similarity without uncertainty-guided distillation. Concretely, ``CLIP-raw'' denotes the default vision encoder used in the LLaVA-1.5 series, i.e., openai/clip-vit-large-patch14-336~\cite{radford2021learning}, where the image--text alignment score is computed directly from the pretrained CLIP representations without our reward calibration procedure. As shown in Table~\ref{tab:clip_raw_appendix}, this baseline performs worse than PSRD on AMBER generative~\cite{amber}, suggesting that raw CLIP similarity alone is insufficient and that the distilled reward model is better aligned with hallucination mitigation.

\begin{table}[t]
  \centering
  \caption{Comparison with a raw CLIP similarity baseline on AMBER generative.}
  \label{tab:clip_raw_appendix}
  \renewcommand{\arraystretch}{1.12}
  \setlength{\tabcolsep}{5pt}
  \begin{tabular}{lcccc}
    \toprule
    Method & CHAIR $\downarrow$ & Cover $\uparrow$ & Hal $\downarrow$ & Cog $\downarrow$ \\
    \midrule
    \textbf{LLaVA-1.5-7B}   & 7.8 & 51.0 & 36.4 & 4.2 \\
    \quad + CLIP-raw & 8.7 & 41.7 & 24.4 & 1.4 \\
    \quad + PSRD     & 3.9 & 48.2 & 20.1 & 2.0 \\
    \bottomrule
  \end{tabular}
\end{table}

\subsection{Comparison Between Natural and Induced Hallucinations}
\label{app:artificial_natural}

To examine whether the hallucinations used for reward-model construction remain representative of naturally occurring errors, we compare induced hallucinations with natural hallucinations generated on clean images.
Specifically, we consider three settings: (1) \emph{Natural}, where hallucinations arise directly from standard generation on clean images; (2) \emph{Induced corruption}, where hallucinations are elicited by our controlled perturbation pipeline; and (3) \emph{Image Gaussian noise}, where the input image is corrupted with Gaussian noise of different strengths $\sigma \in \{25/255, 50/255, 200/255, 500/255\}$.

Table~\ref{tab:artificial_natural_appendix} reports several statistics of the resulting hallucinations, including the average number of hallucinated objects, the unsupported ratio, the semantic drift score, and the noise ratio.
We observe that induced hallucinations remain broadly similar to natural hallucinations in object-grounding behavior, rather than degenerating into random or purely noisy text.
Although stronger image corruption increases the difficulty of grounding and introduces additional noise, the induced errors still preserve a meaningful grounding-related structure.

These results support the validity of our data construction strategy: the induced hallucinations used for training do not merely reflect arbitrary corrupted language patterns, but retain key characteristics of the natural hallucinations that the reward model is expected to detect.

\begin{table*}[t]
  \centering
  \caption{Comparison between natural and induced hallucinations.}
  \label{tab:artificial_natural_appendix}
  \setlength{\tabcolsep}{7pt}
  \renewcommand{\arraystretch}{1.08}
  \begin{tabular}{llcccc}
    \toprule
    Type & Setting & Hallucinated Objects & Unsupported Ratio & Drift & Noise Ratio \\
    \midrule
    Natural & -- & 18.85 & 0.85 & 0.38 & 0.24 \\
    \midrule
    Induced corruption & -- & 22.30 & 0.85 & 0.37 & 0.23 \\
    \midrule
    \multirow{4}{*}{Image Gaussian noise}
      & $\sigma=25/255$  & 18.42 & 0.84 & 0.38 & 0.25 \\
      & $\sigma=50/255$  & 19.22 & 0.85 & 0.37 & 0.25 \\
      & $\sigma=200/255$ & 19.83 & 0.91 & 0.35 & 0.28 \\
      & $\sigma=500/255$ & 20.48 & 0.97 & 0.32 & 0.26 \\
    \bottomrule
  \end{tabular}
\end{table*}

\section{Experiment Details in Section~\ref{sec:experiments}}
\label{supp:exp}
In this section, we provide the details of the evaluation datasets and settings.

\subsection{Details of the Datasets.}
\paragraph{Generative Hallucination Mitigation Task Settings.}
For generative hallucination mitigation tasks, we employ a comprehensive set of metrics across different benchmarks following established protocols~\cite{amber,rohrbach2018object}.
For AMBER, we adopt multiple established metrics including \textbf{CHAIR}~\cite{rohrbach2018object}, \textbf{Cover}~\cite{amber}, \textbf{Hal}~\cite{amber}, and \textbf{Cog}~\cite{amber}.
For Object HalBench: We quantify hallucination using two CHAIR variants: $\mathbf{CHAIR}_i$, the fraction of hallucinated objects in the entire caption; and $\mathbf{CHAIR}_s$, the fraction of captions containing at least one object hallucination.
For MMHalBench, we utilize the \textbf{Hal}~\cite{amber} metric along with the \textbf{Overall} quality score. 
The Overall score is assessed by GPT-4~\cite{gpt4} on a scale of 0 to 6, measuring the overall quality of the generated responses relative to human-generated answers and other ground-truth information of the images.
\paragraph{Discriminative Hallucination Mitigation tasks}
For discriminative hallucination mitigation tasks, we use \textbf{Accuracy} and \textbf{F1-score} to evaluate the performance of LVLMs.
For POPE, this evaluation covers three distinct settings: \textbf{Random}, \textbf{Popular}, and \textbf{Adversarial}. 
The \textbf{ALL} metric represents the arithmetic mean of the scores obtained across these three settings.
\paragraph{Hallucination Classification tasks}
For the evaluation of the reward model distilled by LLaVA1.5-7B, we use the AMBER HalDet and MHal-detect~\cite{gunjal2024detecting} as evaluation datasets.
MHal-detect consists of 16k fine-grained annotations on VQA examples, making it a comprehensive multi-modal hallucination detection dataset for detailed image descriptions.
As a sentence-level binary hallucination classification benchmark, it can be used to evaluate the performance of the reward model.
The AMBER HalDet benchmark is derived from the generative evaluation component of the established AMBER evaluation suite~\cite{amber}.
To construct the dataset, we first collect image captions generated by the LLaVA-1.5-7B. 
These captions were segmented into individual sentences. 
We then apply a strict filtering process: only sentences containing at least two distinct objects were retained, and trivial or semantically empty phrases (e.g., "In this image," or "Additionally,") were removed to ensure content richness.
Each remaining sentence was subsequently evaluated using the official AMBER hallucination assessment tool~\cite{amber} and assigned a binary label: "Hallucinated" or "Non-hallucinated."
Formulated as a binary classification task, AMBER HalDet requires a vision hallucination detector to accurately determine whether a given sentence is consistent with the associated image, thereby serving as a critical measure of fine-grained visual grounding capacity.

\subsection{Details of the Baselines}
\label{sec:appendix-baselines}
For hallucination mitigation tasks, we employ a comprehensive suite of state-of-the-art hallucination mitigation approaches as baselines: 1) \textit{Standard LVLMs}: LLaVA-1.5-7B~\cite{llava1.5} and GPT-4V~\cite{gpt4}; 2) \textit{Fine-tuned LVLMs with externally annotated data}: EOS~\cite{yue2024less}, LLaVA-DPO~\cite{fu2025chip}, HSA-DPO~\cite{xiao2025detecting}, CLIP-DPO~\cite{ouali2024clip}, RLAIF-V~\cite{Yu_2025_CVPR}, HDPO~\cite{fu2025mitigating} and HACL~\cite{jiang2024hallucination}; 3) \textit{Fine-tuned LVLMs via self-improvement}: STIC~\cite{deng2024enhancing} and SENA~\cite{tan2025beyond}. 4) \textit{Post-hoc methods}: VCD~\cite{leng2024mitigating}, ICD~\cite{wang2024mitigating} AVISC~\cite{woo2024don}, M3ID~\cite{favero2024multi}, OPERA~\cite{huang2024opera}, DeCo~\cite{wang2024mllm}, MoD~\cite{chen2025mixture}, ConVis~\cite{park2025convis}, EAZY~\cite{che2025hallucinatoryimagetokenstrainingfree}, ALGA~\cite{an2025mitigating}, ONLY~\cite{wan2025only}, MRGD~\cite{manas2025controlling} and Octopus~\cite{suo2025octopus}. \textit{Among these, M3ID, Octopus, DeCo, MoD and MRGD employ dynamic decoding strategies.}

\subsection{Hyperparameter setting for Efficient Reward-guided Targeted Intervention.}
\label{ref:Hyperparameter_tuning}
We follow the two-stage \emph{Scout-and-Project} design in Efficient Reward-guided Targeted Intervention: (i) a discrete scouting stage that evaluates a small set of top-$k$ candidate decoding branches, and (ii) a projection stage that searches the continuous intervention strength $\alpha$ (up to $\alpha_{\max}$) using a secant-style update with probe step $\delta$, until the reward meets a pre-defined acceptance threshold $\tau = 30$.
Accordingly, we use $k{=}5$, $\delta{=}0.5$, and $\alpha_{\max}{=}3.0$ as the default setting in our experiments, and report ablations in Table~\ref{tab:erti_ablation} with two metrics: CHAIR (hallucination-oriented; lower is better) and Cover (coverage-oriented; higher is better).

\subsection{Details of the Model Evaluation.}
For the generation evaluation of our proposed method in Section~\ref{sec:generalization}, we directly apply the lightweight phase-wise reward model distilled from LLaVA-1.5-7B to LLaVA-Next-7B~\cite{li2024llava} and InstructBlip-7B~\cite{instructblip}, to provide rewards for iteratively dynamic hallucination mitigation. 
This specific cross-model application was motivated by two key observations: Firstly, InstructBlip-7B exhibited a poor intrinsic self-hallucination detection capacity; 
secondly, a reward model independently distilled from LLaVA-Next-7B did not yield substantial performance gains. 
These findings collectively validate the strong cross-model generalization capability of our distilled lightweight reward model.

For discriminative hallucination mitigation tasks, we adopt a rigorous two-stage caption-then-answer evaluation protocol.
In the first stage, the LVLM is prompted to generate a caption for the given image. 
To ensure consistent evaluation, we enforce the response to begin with the phrase, "The image features," irrespective of the input question and without additional instructional guidance. 
The proposed mitigation framework is applied during this generation phase to reduce vision hallucination.
In the second stage, the generated caption is prepended to the original question as augmented contextual information. 
The LVLM is subsequently prompted to generate the final answer based on this integrated input, thereby assessing the downstream discriminative hallucination mitigation tasks.

The reported results for all baselines in hallucination mitigation tasks are sourced directly from their original publications or pertinent cited literature. 
For MRGD~\cite{manas2025controlling}, which utilizes two reward models combined with a weighting coefficient $\gamma$, we report the balanced performance achieved with a coefficient of $\gamma=0.5$.

\subsection{Details for Evaluating Fluency Using LLM}
\label{sec:llm_as_a_judge}
We utilize ChatGPT-4o-mini as a LLM judge to evaluate the fluency and overall quality of the generated content, comparing the outputs of M3ID and our proposed method. The detailed prompt used for this assessment is presented below:

\vspace{+0.3cm}
\hrule height 1pt 
\vspace{+0.1cm}
\noindent \textbf{Prompt: LLM as a judge} 

\begin{quote} 
\small
 Please act as an impartial language evaluator and compare the quality of two text samples written below. Your task is to determine which paragraph demonstrates better fluency, grammar consistency, and overall readability. You should provide a short explanation comparing both texts, focusing on aspects such as word choice, syntax correctness, sentence structure smoothness, and naturalness of expression. Avoid any positional bias --- do not let the order of presentation influence your decision. After your explanation, output your final verdict by strictly following this format:
``[A]'' if sentence A is better, ``[B]'' if sentence B is better, and ``[C]'' if both are equally good. The verdict token must appear on the last line by itself. \\

The Start of Sentence A \\
\textbf{sentence\_A} \\
The End of Sentence A \\
The Start of Sentence B \\
\textbf{sentence\_B } \\
The End of Sentence B 
\end{quote}

 \hrule height 1pt 










\end{document}